\documentclass{article}

     \PassOptionsToPackage{numbers, compress}{natbib}

\usepackage[preprint]{neurips_2022}

\usepackage[utf8]{inputenc} 
\usepackage[T1]{fontenc}    
\usepackage[pagebackref,breaklinks,colorlinks]{hyperref}       
\usepackage{url}            
\usepackage{booktabs}       
\usepackage{amsfonts}       
\usepackage{nicefrac}       
\usepackage{microtype}      
\usepackage{xcolor}         

\usepackage{comment}
\usepackage{algorithmic}

\usepackage{graphicx}
\usepackage{amsmath}
\usepackage{amssymb}
\usepackage{multirow}
\usepackage[american]{babel}
\usepackage[linesnumbered,ruled,vlined]{algorithm2e}
\usepackage{wrapfig}

\usepackage[capitalize]{cleveref}
\crefname{section}{Sec.}{Secs.}
\Crefname{section}{Section}{Sections}
\Crefname{table}{Table}{Tables}
\crefname{table}{Tab.}{Tabs.}

\makeatletter
\DeclareRobustCommand\onedot{\futurelet\@let@token\@onedot}
\def\@onedot{\ifx\@let@token.\else.\null\fi\xspace}

\def\eg{\emph{e.g}\onedot} 
\def\ie{\emph{i.e}\onedot}

\makeatother

\newcommand{\D}{\mathcal{D}}

\newcommand{\OC}{\mathcal{O}}
\newcommand{\w}{\boldsymbol{w}}

\newcommand{\tp}{\mathsf{T}}  
\DeclareMathOperator{\E}{\mathbb{E}}

\usepackage{comment}
\usepackage{subcaption}
\newcommand{\system}{\textsc{FedHM}\xspace}

\makeatletter
\def\algbackskip{\hskip-\ALG@thistlm}
\newcommand{\nosemic}{\renewcommand{\@endalgocfline}{\relax}}
\newcommand{\dosemic}{\renewcommand{\@endalgocfline}{\algocf@endline}}
\let\oldnl\nl
\newcommand{\nonl}{\renewcommand{\nl}{\let\nl\oldnl}}
\makeatother

\newtheorem{assumption}{Assumption}
\newtheorem{lemma}{Lemma}
\newtheorem{theorem}{Theorem}

\newcommand{\rmaxl}{r_{\max,\ell}}

\newcommand{\flow}{f_{\text{low}}}
\newcommand{\bh}{\bar{h}}
\newcommand{\by}{\bar{y}}
\newcommand{\LUV}{L_{\nabla g}}
\newcommand{\diag}{\textnormal{diag}}
\renewcommand{\vec}{\textnormal{vec}}

\title{\system: Efficient Federated Learning for Heterogeneous Models via Low-rank Factorization}

\author{%
  Dezhong Yao, Wanning Pan \\
  Computer Science and Technology \\
  Huazhong University of Science and Technology\\
  \texttt{\{dyao,pwn\}@hust.edu.cn} \\
   \And
  Michael J O'Neill, Yutong Dai \\
  Industrial and Systems Engineering\\
  Lehigh University \\
   \texttt{\{moneill,yud319\}@lehigh.edu} \\
   \AND
  Yao Wan, Hai Jin \\
  Computer Science and Technology \\
  Huazhong University of Science and Technology\\
   \texttt{\{wanyao,hjin\}@hust.edu.cn} \\
   \And
   Lichao Sun \\
   Computer Science and Engineering\\
   Lehigh University \\
   \texttt{lis221@lehigh.edu} \\
}

\begin{document}

\maketitle

\begin{abstract}
One underlying assumption of recent federated learning (FL) paradigms is that all local models usually share the same network architecture and size, which becomes impractical for devices with different hardware resources. A scalable federated learning framework should address the heterogeneity that clients have different computing capacities and communication capabilities. To this end, this paper proposes \system, a novel heterogeneous federated model compression framework, distributing the heterogeneous low-rank models to clients and then aggregating them into a full-rank model. Our solution enables the training of heterogeneous models with varying computational complexities and aggregates them into a single global model. Furthermore, \system significantly reduces the communication cost by using low-rank models. Extensive experimental results demonstrate that \system is superior in the performance and robustness of models with different sizes, compared with state-of-the-art heterogeneous FL methods under various FL settings. 
Additionally, the convergence guarantee of FL for heterogeneous devices is first theoretically analyzed.
\end{abstract}

\section{Introduction}
\label{sec:intro}
Federated Learning (FL)~\cite{mcmahan2016communication,mcmahan2017communication} is a collaborative learning framework that allows each local client to train a global model without sharing local private data.
Due to its privacy-preserving property, it is widely used or deployed on mobile and IoT devices.
Although there are various FL frameworks nowadays, the general FL paradigm consists of the following steps: (1) the server sends the global model to selected clients in each communication round, (2) each selected client trains a local model with its private data, (3) the clients send their trained local models back to the server, and (4) the server aggregates the local models to update the global model, and repeats these steps until the global model converges.

However, the FL paradigm still faces many challenges in practice~\cite{kairouz2019advances,wang2021field}.
One of the urgent challenges is the system heterogeneity that clients in FL may have varying computing capacities and communication capabilities due to hardware differences~\cite{lai2021oort,li2020federated}. It is inefficient to require different clients to train the same model in this heterogeneous setting, since some devices are unable to make full use of their computing resources. 
For example, if a smartphone equipped with a CPU and a cloud server equipped with a GPU participate in FL training, deploying the same models on the two devices will slow down the training process in the cloud server. One straightforward solution is to train heterogeneous models according to the clients' hardware resources and this solution leads to a challenge known as \textit{model heterogeneity}.



Model heterogeneity presents many fundamental challenges to FL, including convergence stability, communication overheads, and model aggregation~\cite{kairouz2019advances}.
HeteroFL~\cite{diao2020heterofl} is the state-of-the-art method that aims to address the heterogeneity from clients that have varying computation capabilities.
HeteroFL applies uniform channel pruning (width slimming) for each local model and aggregates the local models by channels on the server-side.
However, HeteroFL suffers from two major limitations: (1) The small models obtained from naive uniform pruning can not be well-trained in FL. (2) The aggregation strategy ignores the small models' sparse information, and the large models face the risk of over-fitting, especially when the data of the clients are not independent and identically distributed (\ie, Non-IID)~\cite{acar2021federated,karimireddy2020scaffold,li2019fedmd,li2021model,lin2020ensemble,reddi2020adaptive,wang2020federated,zhu2021data}.
Split-Mix~\cite{hong2022efficient} proposes to leverage an ensemble mechanism to address the under-training issue in large models, while the performance of ensemble slimming models still lags far behind the performance of the well-trained large model. 
Additionally, both HeteroFL and Split-Mix aim to improve the performance of the server model (the largest model), while the models deployed on the resources-constrained devices hardly benefit from the heterogeneous training.
Finally, no convergence guarantee is proven for these methods.

To address the aforementioned shortcomings, we propose a federated model compression mechanism for model heterogeneity in FL, \system, which enables multiple devices to train heterogeneous local models with different computing capacities and communication capabilities. 
Our key idea is to compress a large model into small models using low-rank factorization and distribute the compressed models to resource-constrained devices which perform local training. Specifically, \system factorizes deep neural networks (DNNs) into a hybrid network architecture, preserving the model's capacity. 
Subsequently, the server gathers the trained heterogeneous local models from the selected clients and transforms each local model into a full-rank model shape. Then, the global model is easily aggregated from these transformed local models. 
Note that the model factorization is computed on the server, which does not introduce extra computation overhead for the clients.

\label{ref:ctrib}
Overall, the contributions of this paper are as follows. (1) We propose \system, which is the first work that utilizes the low-rank factorization mechanisms of DNNs to solve the model heterogeneity challenge in federated learning. (2) We prove a convergence guarantee for \system in the heterogeneous device setting. (3) Empirically, our proposed method shows superior performance and robustness using small sized models across different scenarios, when compared with state-of-the-art heterogeneous FL methods. (4) Our approach achieves high efficiency in communication and can be flexibly adapted to different heterogeneous scenarios.
\section{Federated Learning on Heterogeneous Devices}
\label{sec:prel}

In typical FL algorithms, each local model $\w_p$ shares the same network architecture with the global model $\w$. The server receives local models and aggregates them into a global model by averaging $\w=\frac{1}{P}\sum^P_{p=1}\w_p$. However, this requires all the clients to train the models with the same structure. This work aims to release the constraint that each local model needs to share the same network architecture as the global model.

\begin{figure*}[t]

    \centering
    \includegraphics[width=0.93\linewidth]{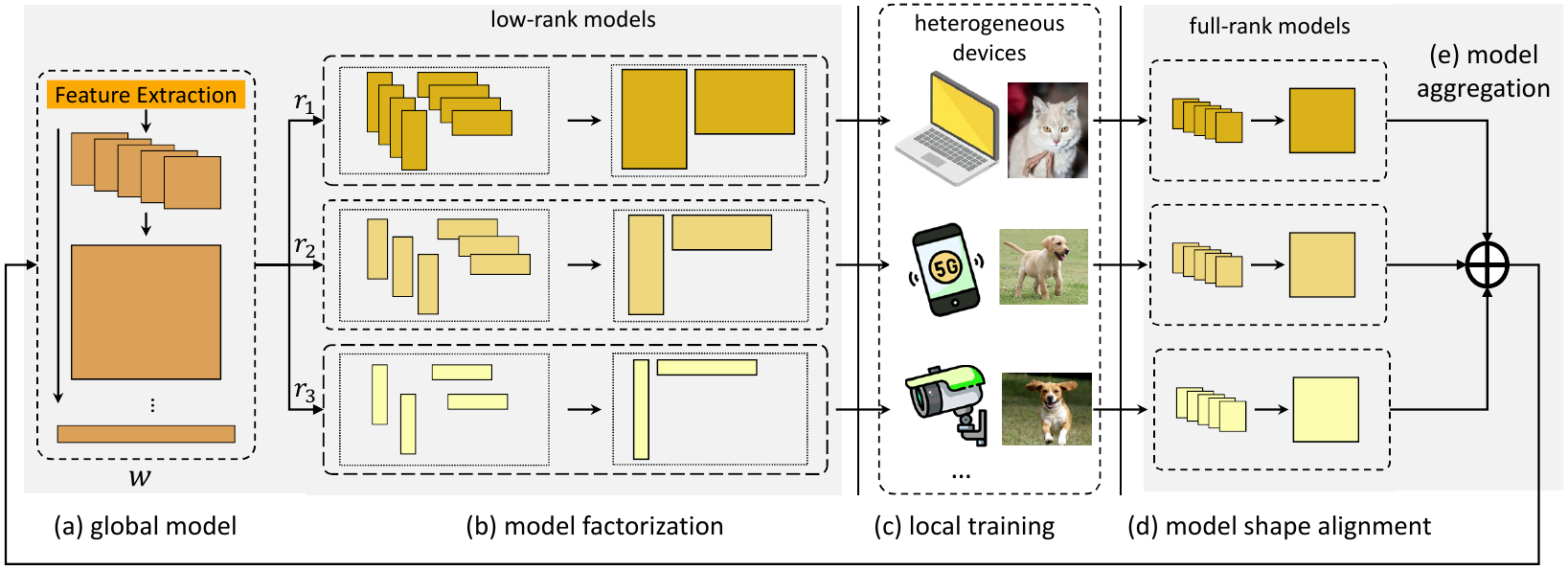}

    \caption{Overview of \system. The global model $\w$ is low-rank factorized to small models and distributed to three clients. The model factorization, model recovery, and model aggregation steps (a), (b), (d), and (e) are in the server side. Only local training step (c) runs in the client side.}
    \label{fig:arch}
\end{figure*}

\label{sec:fedmc}

\subsection{Overview of \system}
In \system, each client is allowed to train a low-rank factorized neural network with a specified size, so that resource-constrained devices can participate in training and devices with strong computing power can make full use of their resources. Before the local training stage, the central server factorizes the global model into different hybrid networks and sends the factorized models to the corresponding participating clients. After receiving the assigned factorized model, the clients train it on their private data with low-rank regularization and send the trained model back to server.

\system consists of three main components, \ie, (1) \textit{local factorized training}, (2) \textit{model shape alignment}, and (3) \textit{model aggregation}. Local factorized training allows clients with heterogeneous hardware resources to train factorized low-rank models with different sizes. This enables each client to optimally utilize its hardware resources. For instance, a cell phone client with $8$G memory capacity can only afford to train a factorized low-rank ResNet-18 with a rank ratio of $\frac{1}{8}$, but a desktop client, on the other hand, with $16$G memory capacity can afford to train a full-rank ResNet-18. The global model $\w$ is factorized to different models and distributed to suitable devices. Upon receiving the factorized models, \system conducts model shape alignment to transform the factorized low-rank models with different sizes back to full-rank shapes, so that each transformed model has an identical shape. Specifically, central model shape alignment transforms factorized models $U\in \mathbb{R}^{m\times r}, V^\top\in \mathbb{R}^{r \times n}$ with different $r$s back to $W\in \mathbb{R}^{m\times n}$. \system finally conducts central aggregation among the transformed models using weighted averaging. An overview of our federated model decomposition is illustrated in Figure~\ref{fig:arch}.


\subsection{Local Factorized Training}

\noindent\textbf{Convolution Layer Factorization.}
For a 2D convolution layer, an input with dimension $(m, H, W)$ convolved with $n$ convolution filters of dimension $(m, k, k)$ to create a $n$-channel output feature map. In our notation, $H, W$ denote the height and width of the input feature map and $k$ denotes the shape of a convolution filter. 
In order to reduce the computational and memory cost, we use the low-rank factorization to modify the architecture of the convolution layers. The core idea is to replace full rank vanilla convolution layers with factorized versions such that the model size and computation complexity are both reduced. 

There is more than one way to adopt factorized convolution (FC) layers. In \system, we utilize the following strategy for convolution layer factorization, which was proposed by~\cite{khodak2020initialization}. For a 2D convolution layer, the parameter matrices is a 4D tensor with dimension $(m, n, k, k)$ where $m, n, k$ denote input channels, output channels, and size of the convolution filters. Instead of factorizing the 4D tensor of a convolution layer directly, we consider factorizing the unrolled 2D matrix. Unrolling the 4D tensor $W$ leads to a 2D matrix with shape $(mk, nk)$. Then factorizing the matrix $W \in \mathbb{R}^{mk\times nk}$ via truncated SVD, leads to $U \in \mathbb{R}^{mk \times r}, V^\top \in \mathbb{R}^{r \times nk}$ with a specific choice of $r$. Reshaping the factorized $U$,$V$ back to 4D tensors leads to $U\in\mathbb{R}^{m \times r \times k \times 1}$, $V^{\top}\in \mathbb{R}^{r \times n \times 1 \times k}$. Therefore, factorizing a convolution layer with dimension $(m, n, k, k)$ returns two convolution layers with dimensions of $(m, r, k, 1)$ (a convolution layer with $r$ $(k \times 1)$ convolution filters) and $(r, n, 1, k)$ (a convolution layer with $n$ $(1 \times k)$ convolution filters). An alternative approach to our proposed low-rank factorization is to directly conduct tensor decomposition, \eg, the CP ~\cite{lebedev2015speeding} and Tucker~\cite{phan2020stable} decomposition to directly factorize the 4D tensor weights. In this work, we do not consider tensor decomposition, as they were observed to have lower accuracy and more computational cost than low-rank factorization in~\cite{khodak2020initialization}.

\noindent\textbf{Hybrid Network Architecture.}
Since the low-rank factorized network is an approximation of the original network, 
an approximation error will be introduced when factorizing each layer. Approximation errors in the early layers can be accumulated and propagated to the later layers, which will significantly impact the model accuracy~\cite{wang2021pufferfish}. A straightforward idea is to only factorize the later layers to preserve the final model accuracy. Moreover, especially for CNNs, the number of parameters in later layers dominates the entire network size, \eg, the last $3$ bottleneck residual blocks of ResNet-50 contains $\sim 60\%$ of the entire model parameters. Thus, factorizing the later layers does not sacrifice the model compression rate. 
To achieve this goal, we utilize the hybrid $[\rho,L,\gamma]$ network architecture, \ie, $\w^\gamma=\{W_1,W_2,\ldots,W_{\rho},U_{\rho+1}, V^\top_{\rho+1},\ldots, U_L, V^\top_L\}$, where $\rho$ denotes the number of layers are \textit{not} factorized, $L$ is the total number of layers, and $\gamma = rank(U_\ell V^\top_\ell)/rank(W_\ell)$ for all $\ell\in \{\rho+1, \cdots, L\}$. We remark that $\rho$ serves as  a hyper-parameter that balances the model compression ratio and the final model accuracy.

\noindent\textbf{Initialization and Regularization of Factorized Layers.}
As illustrated in~\cite{khodak2020initialization}, the performance of the low-rank models can be improved through initializing with spectral initialization (SI) and regularizing with Frobenius decay (FD). The Frobenius decay penalty on matrices $U$, $V^{\top}$ is $\frac{\lambda}{2} \left\|UV^{\tp}\right\|^2_F$ where $\lambda$ is the coefficient of regularization which controls the strength of the penalty. Spectral initialization is naturally implemented during the FL model distribution process (by using the SVD to decompose the server model) and we use Frobenius decay for training the low-rank layers, so that the low-rank layers are appropriately initialized and regularized.

\subsection{Model Shape Alignment and Aggregation}

At round $t$, after the server receives the local update from the $p$-th client, which is denoted as  $\w_{p,t}^{\gamma_p}$ where $\gamma_p$ is participant $p$'s rank ratio, \system conducts model shape alignment operation, to recover the full-size model $\w_{p,t}$ as
\begin{equation}\label{eq:shape.alignment}
    \w_{p,t+1} = \{W_1,W_2,\ldots,W_{\rho},U_{\rho+1}V^\top_{\rho+1},\ldots, U_LV^\top_L\}.
\end{equation}
We drop the subscripts on $p$ and $t$ to avoid cluttered notation on the right hand side of Eq.~\eqref{eq:shape.alignment}.

As the number of parameters represents the knowledge of each model, we use a weighted aggregation scheme to derive the global model from the local updates as 
\begin{equation}\label{eq:fedmc_agg}
\begin{aligned}
  \w_{t+1} & = \sum_{p=1}^P \alpha_p\w_{p,t+1}\,, \quad
   \alpha_p & = \frac{\exp{\frac{\gamma_p}{\tau}}}{\sum^{P}_{p=1} \exp{\frac{\gamma_p}{\tau}}}\,,
\end{aligned}
\end{equation}
where $P$ is the number of active clients and $\tau>0$ denotes the softmax temperature. Due to space constraints, the proposed \system is sketched in Algorithm~\ref{alg:fedhms} and a more precise description can be found in the Appendix.

\begin{algorithm}[t]

\caption{\system: Federated Learning for Heterogeneous Models. (Sketch)}
\label{alg:fedhms}

  \SetKwInOut{KwIn}{Input} \SetKwInOut{KwOut}{Output}
  \KwIn{Dataset $[\D_p]_{p=1}^P$ distributed on $P$ clients, total communication rounds $T$, local epoch length $E$, a set of rank ratios $\{\gamma_i\}_{i=1}^{\Gamma}$, the number of un-factorized layers $\rho$.}
  \KwOut{Final model $\w_T$ }
\nosemic\nonl \textbf{ServerExecute:} \tcp*[h]{server side} \;
  Initialize $L$-layer network with weight $\w_0=\{W_1, \ldots, W_{L}\}$\;
  \For{$t=0,\ldots,T-1$}{
        \For{$i=1,\ldots,\Gamma$}
        { 
            Factorize the network $\w_t$ to a $[\rho, L, \gamma_i]$ hybrid network $\w_t^{\gamma_i}$.
        }
        \ForPar{each client $p$}{
         Based on the computation capability, choose the hybrid network $\w_{p,t}^{\gamma_p}$ from $\{\w_{t}^{\gamma_i}\}_{i=1}^{\Gamma}$.\;
            $\w_{p,t+1}^{\gamma_p}\leftarrow$ \textbf{LocalUpdate}($\w_{p,t}^{\gamma_p}$).\;
        Perform model shape alignment on the hybrid network $\w_{p,t+1}^{\gamma_p}$ to get $\w_{p,t+1}$ as in Eq.~\eqref{eq:shape.alignment}.\;
        }
        Aggregate the local updates from clients to get a new network weight $\w_{t+1}$ as Eq.~\eqref{eq:fedmc_agg}.
}
\Return $\w_T$
\BlankLine
\nosemic\nonl \textbf{LocalUpdate}($\w$): \tcp*[h]{client side} \;
    Train the hybrid network for $E$ epochs and regularize with Frobenius decay.\;
    \Return local solution to server.\;
\end{algorithm}

\section{Theoretical Analysis}

\subsection{Computational Complexity and Model Size}

\begin{table}[!htb]

\caption{Comparison of parameters and computational complexity for full-rank and low-rank FC and Conv. layers.}
\label{tbl:complex_compare}
\renewcommand\arraystretch{1.0}
\centering
\resizebox{.7\linewidth}{!}{
\begin{tabular}{lcc}
\toprule
Layer &   \# Params.  & Comp. Complexity   \\
\midrule
Vanilla FC &  $mn$ & $\OC(mn)$ \\
Factorized FC & $r(m+n)$  & $\OC(r(m+n))$ \\
\midrule
Vanilla Conv. & $mnk^2$  & $\OC(mnk^2 H W)$ \\
Factorized Conv. & $rk(m+n) $  & $rkHW(m+n)$ \\
Factorized Conv. in~\cite{wang2021pufferfish} & $r(mk^2+n)$  & $rHW(mk^2 + n)$ \\
\bottomrule
\end{tabular}
}
\end{table}

We summarize the computational complexity and the number of parameters in the vanilla and low-rank FC, convolution layers in Table~\ref{tbl:complex_compare}. For the model sizes of both types of FC layers, low-rank factorization reduces the model dimension from $(m, n)$ to $(m, r), (r, n)$. For the computational complexity of the FC layers, it is also easy to see that $xW$ requires $\mathcal{O}(mn)$ and $xUV^\top$ requires $\mathcal{O}(mr+rn)=\mathcal{O}(r(m+n))$. For a convolution layer, the un-factorized layer has dimension $(m, n, k, k)$ while the factorized low-rank layer has dimensions $(m, r, k, 1), (r, n, 1, k)$ as discussed before. Thus, the factorized convolution reduces the total number of parameters from  $mnk^2$ to $rmk+rnk$. For the computational complexity of convolution layers, an un-factorized convolution layer requires $\mathcal{O}(mnk^2HW)$ as there are $n$ filters with size $m\times k\times k$, and the filter has to convolve over $H\times W$ pixels (controlled by a stride size). Using the same analysis, one can easily see that the factorized low-rank convolution layer needs $\mathcal{O}(mrkHW+nrkHW)$ operations to convolve a input with $HW$ pixels and $m$ input channel. We also compare our convolution layer factorization strategy against another popular factorization strategy used in Pufferfish~\cite{wang2021pufferfish} (shown in Table~\ref{tbl:complex_compare}). One can see that for both model size and computational complexity, our factorization strategy balances the $k$ term between the input dimension $m$ and output dimension $n$ while the strategy used by ~\cite{wang2021pufferfish} places all $k^2$ terms in the input dimension $m$. Thus, unless the input dimension is much smaller than output dimension, our strategy usually leads to smaller model size and lower computational complexity. A low-rank factorized network enjoys a smaller network of parameters and lower computational complexity which is indicated in Table~\ref{tbl:complex_compare}. We note that both the computation and communication efficiencies are improved, as the communication cost is proportional to the number of parameters.

\subsection{Convergence Analysis}
\label{sec:cnvg}

In this section, we provide a convergence result for Algorithm \ref{alg:fedhms} under the assumption of full client participation. We consider the minimization of the loss function $f$:
\begin{equation} \label{eq:fdefmain}
    \underset{w}{\min} \; f(w) = \frac{1}{P} \sum_{p=1}^P f_p(w), \text{ and } f_p(w) := \E_{\zeta_p \sim \D_p}[F_p(w;\zeta_p)],
\end{equation}
 where $f_p$ are smooth non-convex functions and $\D_p$ can vary per participant. To prove our convergence result, we need two assumptions, the first of which is standard in the federated learning literature.
\begin{assumption} (Informal) \label{assum:informal1}
    Each $f_p$ is continuously differentiable with Lipschitz continuous gradients and lower bounded by $\flow \in \mathbb{R}$. In addition, the stochastic gradients are unbiased estimators of the true gradient and the second moment of the stochastic gradients is bounded (with constant $\kappa_g$).
\end{assumption}
In addition to this assumption, we make the following assumption that is specific to the low-rank factorization strategy of Algorithm \ref{alg:fedhms}.
\begin{assumption} (Informal) \label{assum:informal2}
    For each iteration and participant, the components $U$ and $V$ of the factorized layers are bounded in norm. In addition, at every aggregation step, the difference between the full model and low-rank models is bounded in norm (by $\kappa_{\sigma}$).
\end{assumption}
With these assumptions, we can present our convergence result.
\begin{theorem} (Informal) \label{thm:convergencemain}
    Let Assumptions \ref{assum:informal1} and \ref{assum:informal2} hold. Let $g$ be the loss function applied to the factorized network. Then,
    \begin{equation*}
        \frac{1}{T E} \sum_i \E[\|\nabla g(\hat{h}_i)\|^2] \leq \OC\left(\frac{f(w_0) - \flow}{T E \eta} + \frac{\kappa_g \eta }{P} + \kappa_g \max\{E \eta, E^2 \eta^2\} + \kappa_{\sigma}\left(1 + \frac{1}{E \eta}\right) \right),
    \end{equation*}
    where $\eta$ is the learning rate and $\hat{h}_i$ is the average of the iterates at time $i$.
\end{theorem}
The first three terms essentially match those found in standard results in the literature \cite{yu2019parallel}. The final term, however, is unique to the heterogeneous device setting and is a direct consequence of the different capabilities of the participants in the network.

\section{Experiments and Analysis}
\label{sec:epx}

\subsection{Experimental Setup}
\noindent\textbf{Datasets and Model Setting.}
To validate the effectiveness of our proposed approach, we adopt three datasets (\ie, CIFAR-10, CIFAR-100~\cite{krizhevsky2009learning} and Tiny-ImageNet\footnote{ \url{https://www.kaggle.com/c/tiny-imagenet}}) that have been widely used in the area of federated computer vision.
According to the complexity of the datasets, we conduct experiments on CIFAR-10, CIFAR-100 and Tiny-ImageNet using ResNet-18, ResNet-34 and ResNet-34, respectively.
To adapt to the FL setting, we set the number of clients to 20, and we sample $50\%$ of the clients at each round to train models $10$ local epochs. 

 
\noindent\textbf{Baselines.}
We compare \system with the following baselines, including (1) \textbf{FedAvg}~\cite{mcmahan2017communication}, which trains the slimming model on all clients, (2) \textbf{HeteroFL}~\cite{diao2020heterofl}, which leverages the network slimming strategy to compress the heterogeneous client models and aggregate them by channels, and (3) \textbf{Split-Mix}~\cite{hong2022efficient}, which ensembles several slimming models to provide heterogeneous models.

\noindent\textbf{System Heterogeneity.}
The system heterogeneity in our experiments is pre-defined as different levels of computational complexity, following~\cite{diao2020heterofl}. We annotate the \textbf{fixed heterogeneous setting} for clients with a fixed assignment of computation complexity levels, and annotate the \textbf{dynamic heterogeneous setting} for clients by uniformly sampling computation complexity levels at each communication round. For HeteroFL, we adjust the shrinkage ratio to vary the model size and extend two variants based on HeteroFL (named HeteroFL-V2 and HeteroFL-V3) by changing the heterogeneous model groups to better study the impact of different model combinations. As for Split-Mix, we control the model size by modifying the number of base models.
For \system, we set a fixed parameter $\rho$ and construct the computation complexity of \system by varying the rank ratio $\gamma$. Under the dynamic heterogeneous setting, the larger models should be given higher weight since the models are trained on all the clients, we set the softmax temperature $\tau=5$. For the fixed heterogeneous setting, we set $\tau=+\infty$ to equally aggregate the models. For a more detailed discussion of hyper-parameters, please refer to the Appendix.





\noindent\textbf{Data Heterogeneity.}
Following prior work~\cite{hsu2019measuring}, we sample the disjoint Non-IID client data using the Dirichlet distribution $\bf{Dir}(\alpha)$, where $\alpha$ denotes the concentration parameter. We set the $\alpha$ to $0.5$.


 

\subsection{Performance Evaluation}\label{sec:exp_perf}
We evaluate the Top-1 accuracy of models with different sizes in the IID and Non-IID settings under the dynamic and fixed heterogeneous FL settings.

\begin{table*}[tb]
\vspace{-0.8em}
\caption{Top-1 test accuracy comparison with different methods under dynamic heterogeneous setting. }
\label{tbl:dynamic_hete}
\resizebox{1.\linewidth}{!}{
\begin{tabular}{l|l|cccccccccccc}
\hline
\multirow{2}{*}{DataSet}   & \multirow{2}{*}{Setting} & \multicolumn{2}{c}{Split-Mix} & \multicolumn{2}{c}{HeteroFL} & \multicolumn{2}{c}{HeteroFL-V2} & \multicolumn{2}{c}{HeteroFL-V3} & \multicolumn{2}{c}{FedAvg} & \multicolumn{2}{c}{\system}        \\ \cline{3-14} 
                           &                          & \#Params     & Acc & \#Params      & Acc      & \#Params      & Acc      & \#Params      & Accuracy      & \#Params        & Acc       & \#Params      & Acc       \\ \hline
\multirow{8}{*}{CIFAR-10}  & \multirow{4}{*}{IID}      & 12.33M     & 90.34  & 11.17M          & 93.19         & 4.29M($\times$2)          & 93.19         & 2.80M($\times$3)           & 93.18              & 1.37M($\times$4)             & 92.66          & \textbf{11.17M} & \textbf{93.50} \\
                           &                           & 4.12M     & 89.87 & 4.29M           & 93.13         & 2.80M           & 92.77         & 1.37M           & 92.14              &                   &                & \textbf{4.16M}  & \textbf{93.52} \\
                           &                           & 2.75M     & 89.44 & 2.80M           & 92.93         & 1.37M           & 92.01         &                 &               &                   &                & \textbf{2.21M}  & \textbf{93.49} \\
                           &                           & 1.37M     & 88.13 & 1.37M           & 92.16         &                 &               &                 &               &                   &                & \textbf{1.24M}  & \textbf{93.43} \\ \cline{2-14} 
                           & \multirow{4}{*}{Non-IID}  & 12.33M     & 85.93 & 11.17M          & 91.57         & 4.29M($\times$2)           & 91.14         & 2.80M($\times$3)           & 91.05         & 1.37M($\times$4)             & 90.21          & \textbf{11.17M} & \textbf{91.59} \\
                           &                           & 4.12M     & 84.82 & 4.29M           & \bf{91.53}         & 2.80M           & 91.03         & 1.37M           & 89.93         &                   &                & \textbf{4.16M}  & 91.51 \\
                           &                           & 2.75M     & 84.54 & 2.80M           & 91.03         & 1.37M           & 89.52         &                 &               &                   &                & \textbf{2.21M}  & \textbf{91.55} \\
                           &                           & 1.37M     & 83.27 & 1.37M           & 90.19         &                 &               &                 &               &                   &                & \textbf{1.24M}  & \textbf{91.47} \\ \hline
\multirow{8}{*}{CIFAR-100} & \multirow{4}{*}{IID}      & 23.97M     & 69.38 & 21.33M          & 66.56         & 8.77M($\times$2)          & 67.32         & 5.35M($\times$3)           & 66.81         & 3.42M($\times$4)             & 63.07          & \textbf{21.33M}  & \textbf{72.38} \\
                           &                           & 10.22M     & 68.26 & 8.77M           & 66.39         & 5.35M           & 66.32         & 3.42M           & 65.31         &                   &                & \textbf{8.40M}  & \textbf{72.45} \\
                           &                           & 6.81M     & 67.63 & 5.35M           & 65.68         & 3.42M           & 65.17         &                 &               &                   &                & \textbf{4.99M}  & \textbf{72.44} \\
                           &                           & 3.42M     & 62.53 & 3.42M           & 64.11         &                 &               &                 &               &                   &                & \textbf{3.27M}  & \textbf{72.29} \\ \cline{2-14} 
                           & \multirow{4}{*}{Non-IID}  & 23.97M     & 66.98 & 21.33M          & 67.37         & 8.77M($\times$2)           & 66.76         & 5.35M($\times$3)            & 67.25             & 3.42M($\times$4)             & 66.69          & \textbf{21.33M}  & \textbf{69.05} \\
                           &                           & 10.22M     & 65.81 & 8.77M           & 67.30         & 5.35M           & 66.04         & 3.42M           & 65.25             &                   &                & \textbf{8.40M}  & \textbf{69.07} \\
                           &                           & 6.81M     & 64.61 & 5.35M           & 66.26         & 3.42M           & 64.75         &                 &               &                   &                & \textbf{4.99M}  & \textbf{69.03} \\
                           &                           & 3.42M     & 60.17 & 3.42M           & 64.56         &                 &               &                 &               &                   &                & \textbf{3.27M}  & \textbf{68.90} \\ \hline
\multirow{8}{*}{Tiny-ImageNet}& \multirow{4}{*}{IID}  & 24.11M     & 55.88 & 21.38M   &  54.63        & 8.80M($\times$2)          & 53.96        & 5.38M($\times$3)   & 53.45              & 3.44M($\times$4)       & 51.59          & \textbf{21.38M} & \textbf{57.80} \\
                           &                           & 10.33M     & 54.35 & 8.80M      & 54.54    & 5.38M           & 52.71         & 3.44M           & 51.27              &                   &                & \textbf{8.45M}  & \textbf{57.77} \\
                           &                           & 6.89M     & 52.24 & 5.38M      & 52.97      & 3.44M           & 50.60         &                 &               &                   &                & \textbf{5.04M}  & \textbf{57.82} \\
                           &                           & 3.44M     & 48.99 & 3.44M      &51.59          &                 &               &                 &               &                   &                & \textbf{3.33M}  & \textbf{57.46} \\ 
                           \cline{2-14} & \multirow{4}{*}{Non-IID}  & 24.11M     & \bf{53.87} & 21.38M  &52.88     & 8.80M($\times$2) & 53.00        & 5.38M($\times$3)& 52.23         & 3.44M($\times$4)        & 50.19          & \textbf{21.38M} & 52.38 \\
                           &                           & 10.33M     & 52.07 & 8.80M   &\bf{52.57}     & 5.38M           & 51.59         & 3.44M           & 49.69         &                   &                & \textbf{8.45M}  & 52.37 \\
                           &                           & 6.89M     & 50.35 & 5.38M   &51.55     & 3.44M           & 49.42         &                 &               &                   &                & \textbf{5.04M}  & \bf{52.35} \\
                           &                           & 3.44M     & 47.34 & 3.44M   &49.35     &                 &               &                 &               &                   &                & \textbf{3.33M}  & \bf{51.75} \\ \hline
                         
\end{tabular}
}

\end{table*}

\begin{table*}[tb]
\caption{Top-1 test accuracy comparison with different methods under fixed heterogeneous setting.}
\label{tbl:HeteroFL_fix_heter}
\resizebox{1.\linewidth}{!}{
\begin{tabular}{l|l|cccccccccccc}
\hline
\multirow{2}{*}{DataSet}  & \multirow{2}{*}{Setting} & \multicolumn{2}{c}{Split-Mix} & \multicolumn{2}{c}{HeteroFL} & \multicolumn{2}{c}{HeteroFL-V2} & \multicolumn{2}{c}{HeteroFL-V3} & \multicolumn{2}{c}{FedAvg} &\multicolumn{2}{c}{\system}   \\ \cline{3-14} 
                          &                          & \#Params     & Acc & \#Params     & Acc    & \#Params      & Acc      & \#Params      & Acc      & \#Params & Acc  & \#Params      & Acc     \\ \hline
\multirow{8}{*}{CIFAR-10} & \multirow{4}{*}{IID}     &  12.33M    & 90.34 & 11.17M         & 89.82       & 4.29M($\times$2)           & 91.14         & 2.80M($\times$3)          & 91.83  & 1.37M($\times$4) &  92.66  & \bf{11.17M}     & \textbf{93.46} \\
                          &                          & 4.12M     & 89.92 & 4.29M          & 89.97       & 2.80M           & 91.03         & 1.37M           & 90.15    & &     & \bf{4.16M}      & \textbf{93.37} \\
                          &                          & 2.75M     & 89.33 & 2.80M          & 89.38       & 1.37M           & 89.52         &                 &         &&      & \bf{2.21M}      & \textbf{93.40} \\
                          &                          & 1.37M     & 88.30 & 1.37M          & 87.83       &                 &               &                 &        &&       &\bf{1.24M}      & \textbf{93.40} \\ \cline{2-14} 
                          & \multirow{4}{*}{Non-IID} & 12.33M     & 85.91 & 11.17M         & 84.49       & 4.29M($\times$2)   &  87.61         & 2.80M($\times$3)       &  87.46     &1.37M($\times$4)&  90.21    & \bf{11.17M}     & \bf{91.22}          \\
                          &                          & 4.12M     & 84.85 & 4.29M          & 86.44       & 2.80M           & 86.93         & 1.37M               & 84.45   &&          & \bf{4.16M}      & \bf{91.21}          \\
                          &                          & 2.75M     & 84.48 & 2.80M          & 86.28       & 1.37M           &  84.20            &                 &      &&         & \bf{2.21M}      & \bf{91.23}          \\
                          &                          & 1.37M     & 83.30 & 1.37M          & 84.08       &                 &               &                 &      &&         & \bf{1.24M}      & 
                          \bf{91.17}          \\ \hline
\multirow{8}{*}{CIFAR-100} & \multirow{4}{*}{IID}     & 23.97M     & 69.42& 21.33M         & 56.01       & 8.77M($\times$2)           & 62.05         & 5.35M($\times$3)          & 64.24         & 3.42M($\times$4)                          & 63.07                        & \bf{21.33M}     & \textbf{70.23} \\
                           &                          & 10.22M     & 68.30 & 8.77M          & 59.23       & 5.35M           & 61.83         & 3.42M           & 61.52         &                                &                              & \bf{8.40M}      & \textbf{70.23} \\
                           &                          & 6.81M     & 67.60 & 5.35M          & 59.93       & 3.42M           & 59.20         &                 &               &                                &                              & \bf{4.99M}      & \textbf{70.26} \\
                           &                          & 3.42M     & 62.59 & 3.42M          & 57.95       &                 &               &                 &               &                                &                              & \bf{3.27M}      & \textbf{70.00} \\ \cline{2-14} 
                           & \multirow{4}{*}{Non-IID} & 23.97M     & 67.02 & 21.33M         & 53.55       & 8.77M($\times$2)           & 60.42         & 5.35M($\times$3)           & 64.37         & 3.42M($\times$4)                          & 66.69                        & \bf{21.33M}      & \bf{69.04} \\
                           &                          & 10.22M      & 65.84 & 8.77M          & 58.40       & 5.35M           & 59.16         & 3.42M           & 60.06         &                                &                              & \bf{8.40M}      & \bf{68.91} \\
                           &                          & 6.81M     & 64.65 & 5.35M          & 58.61       & 3.42M           & 57.67         &                 &               &                                &                              & \bf{4.99M}      & \bf{68.97} \\
                           &                          & 3.42M     & 60.24 & 3.42M          & 57.20       &                 &               &                 &               &                                &                              & \bf{3.27M}      & \bf{68.86} \\ \hline
\multirow{8}{*}{Tiny-ImageNet}     & \multirow{4}{*}{IID}  & 24.11M     & 55.78   & 21.38M          & 49.54         & 8.80M($\times$2)          & 50.15       & 5.38M($\times$3)  & 50.79    & 3.44M($\times$4)       & 51.59          & \textbf{21.38M} & \textbf{57.72} \\
                           &                          & 10.33M     &  54.69  & 8.80M           & 49.90         & 5.38M           & 48.08         & 3.44M           & 46.34         &                   &                & \textbf{8.45M} & \textbf{57.69} \\
                           &                          & 6.89M     &  52.81   & 5.38M           & 48.98        & 3.44M           & 45.53         &                 &               &                   &                & \textbf{5.04M}  & \textbf{57.67} \\
                           &                          & 3.44M     & 48.86   & 3.44M           & 46.52         &                 &               &                 &               &                   &                & \textbf{3.33M}  & \textbf{57.49} \\ \cline{2-14} 
                           & \multirow{4}{*}{Non-IID} & 24.11M     & \textbf{53.79}  & 21.38M          & 45.19         & 8.80M($\times$2)           &46.98         & 5.38M($\times$3)            & 49.19           & 3.44M($\times$4)             & 50.19          & \textbf{21.38M}  & 52.44 \\
                           &                          & 10.33M     & 52.27  & 8.80M           & 46.14         & 5.38M           & 46.21         & 3.44M           & 44.33             &                   &                & \textbf{8.45M}  & \textbf{52.48} \\
                           &                          & 6.89M     & 50.76   & 5.38M           & 45.12         & 3.44M           & 42.68         &                 &               &                   &                &\textbf{5.04M}   & \textbf{52.48} \\
                           &                          & 3.44M     & 47.04   & 3.44M           & 43.19        &                 &               &                 &               &                   &                & \textbf{3.33M}  & \textbf{52.35} \\ \hline
\end{tabular}
}
\end{table*}


\noindent\textbf{Performance under Dynamic Heterogeneous Setting.}
Table~\ref{tbl:dynamic_hete} shows the performance of \system and baselines in terms of Top-1 test accuracy under the dynamic heterogeneous setting. 
From this table, it is clear to see that both the server and client models of \system consistently outperform the baselines on CIFAR datasets.
When comparing with FedAvg, we find that both \system and HeteroFL improve the performance of the server model, while Split-Mix does not beat FedAvg in all of the scenarios under the dynamic setting.
When comparing \system with different variants of HeteroFL and Split-Mix, we observe that \system shows superiority in terms of the accuracy of the smallest model. 
On CIFAR-100, \system surpasses HeteroFL by 5.90\% and 1.09\% in terms of Top-1 accuracy, under the IID and Non-IID settings respectively. 
The improvements for small models are helpful for dynamically changing computational capability.
Furthermore, we also find that the performance gain achieved by \system compared with baselines becomes larger when switching the dataset from CIFAR-10 to datasets with more complexity, \ie, CIFAR-100 and Tiny-ImageNet. 
We attribute this to direct width slimming on larger model leading to a significant performance drop when compared to the original model. In contrast, the small models in \system benefit from the hybrid network structure and weighted aggregation strategy, which leads to improved performance.


\noindent\textbf{Performance under Fixed Heterogeneous Setting.}
Here we conduct experiments that are same to those have been done in the fixed heterogeneous setting. As shown in Table~\ref{tbl:HeteroFL_fix_heter}, the performance of HeteroFL drops under the fixed heterogeneous setting, while \system and Split-Mix are robust. 
Comparing the performance of the server model, \system outperforms Split-Mix 3.12\%/5.31\% and 0.81\%/2.02\% in terms of Top-1 test accuracy in the CIFAR10 IID/Non-IID, CIFAR100 IID/Non-IID settings. 
Comparing the performance of the smallest model, \system outperforms all of the baselines in terms of test accuracy while using fewer parameters.
We find that Split-Mix consistently outperforms HeteroFL under the fixed heterogeneous setting, which can be attributed to the fact that large models in HeteroFL are only trained on strong devices, facing the risk of over-fitting. \system reduces over-fitting by aggregating the sparse information of the small models, while the sparse information is ignored in HeteroFL.
Though Split-Mix aims to address the over-fitting issue by ensembling the smallest models, it does not improve the performance of the smallest model. \system reduces the risk of large model over-fitting by aggregating with transformed low-rank models. Additionally, the low-rank models also benefit from aggregating with the larger models. 

\subsection{Efficiency Analysis}\label{sec:exp_effi}
\begin{figure}[tb]

\centering
\begin{subfigure}{0.325\columnwidth}

\includegraphics[width=\linewidth]{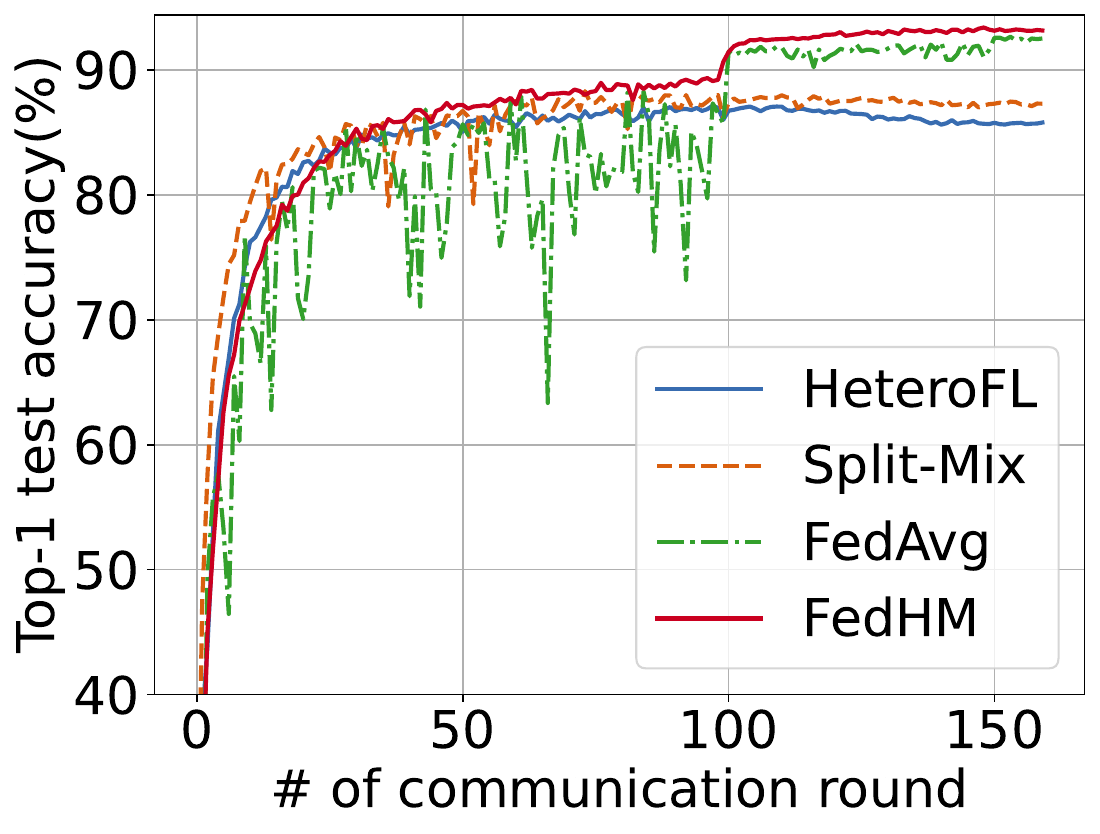} 
\caption{CIFAR-10.}
\end{subfigure}
\label{fig:hete_cifar10}
\begin{subfigure}{0.325\columnwidth}
\includegraphics[width=\columnwidth]{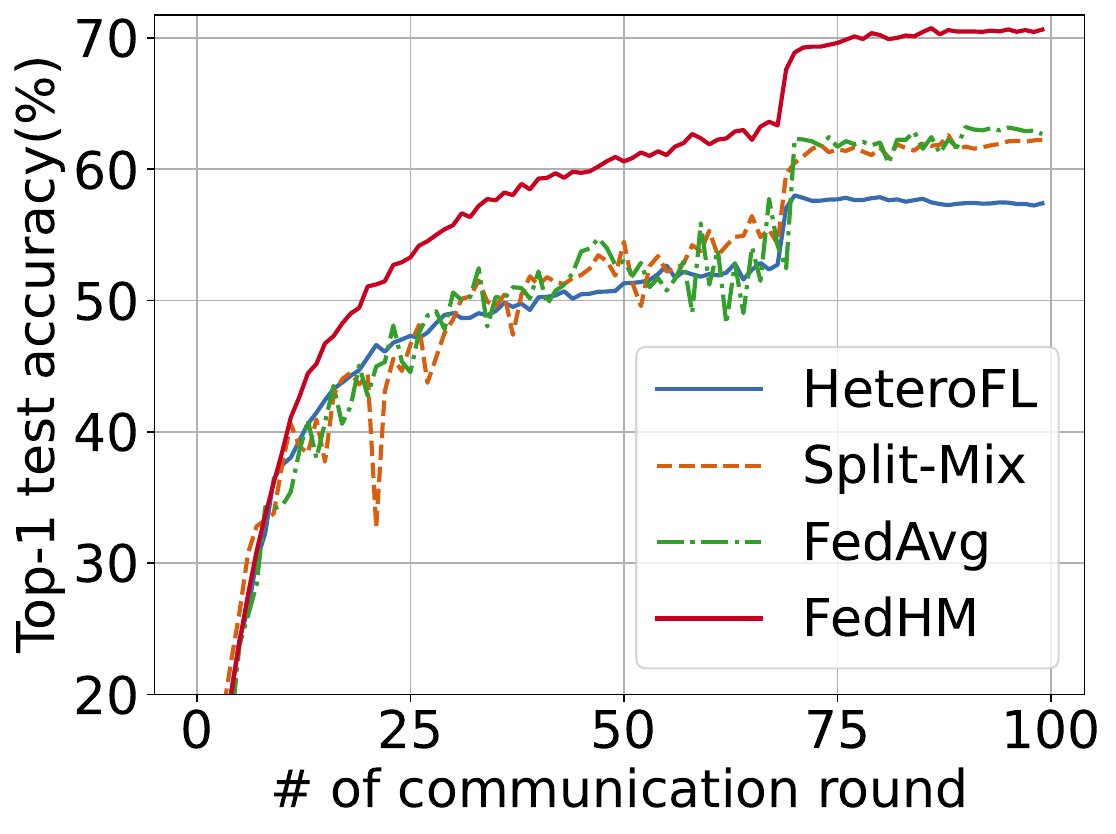}
\caption{CIFAR-100.}
\end{subfigure}
\begin{subfigure}{0.325\columnwidth}
\includegraphics[width=\columnwidth]{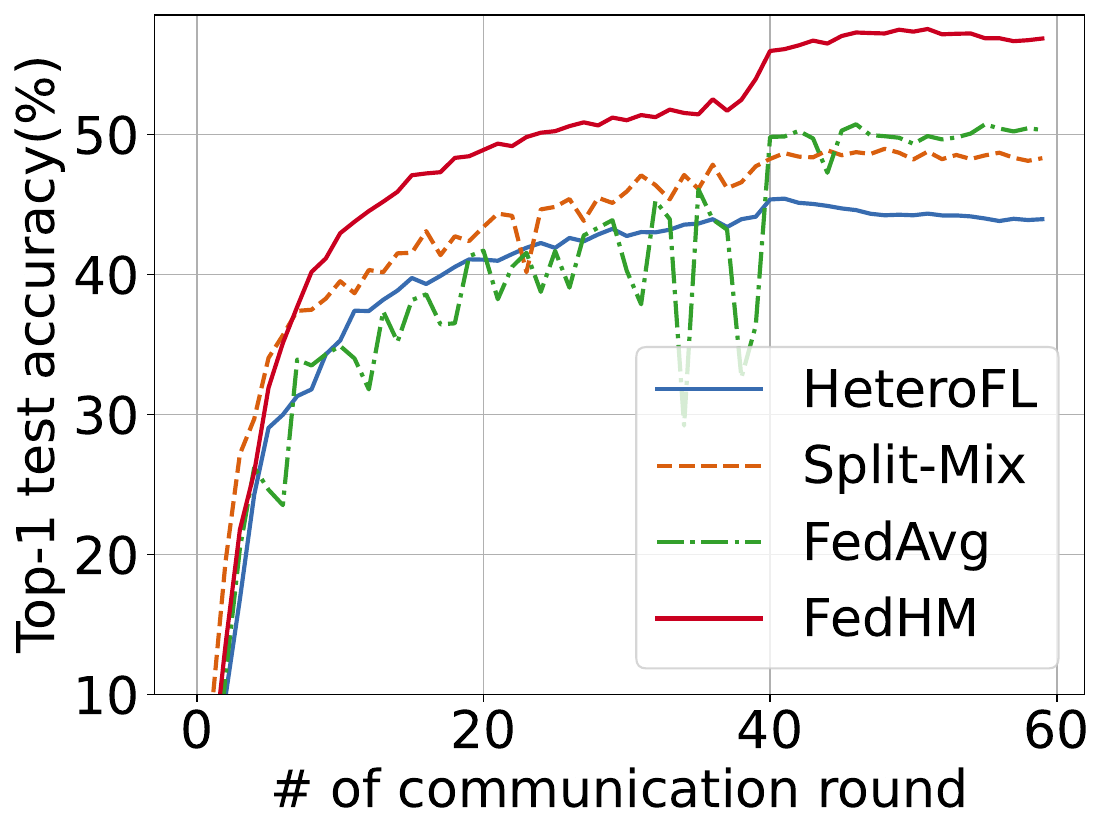}
\caption{Tiny-ImageNet.}
\end{subfigure}

\caption{Evaluating different methods in terms of convergence speed of the smallest model.}
\label{fig:hete_iid}

\end{figure}

\begin{figure}[tb]
\centering
\begin{subfigure}{0.245\textwidth}
\includegraphics[width=.99\columnwidth]{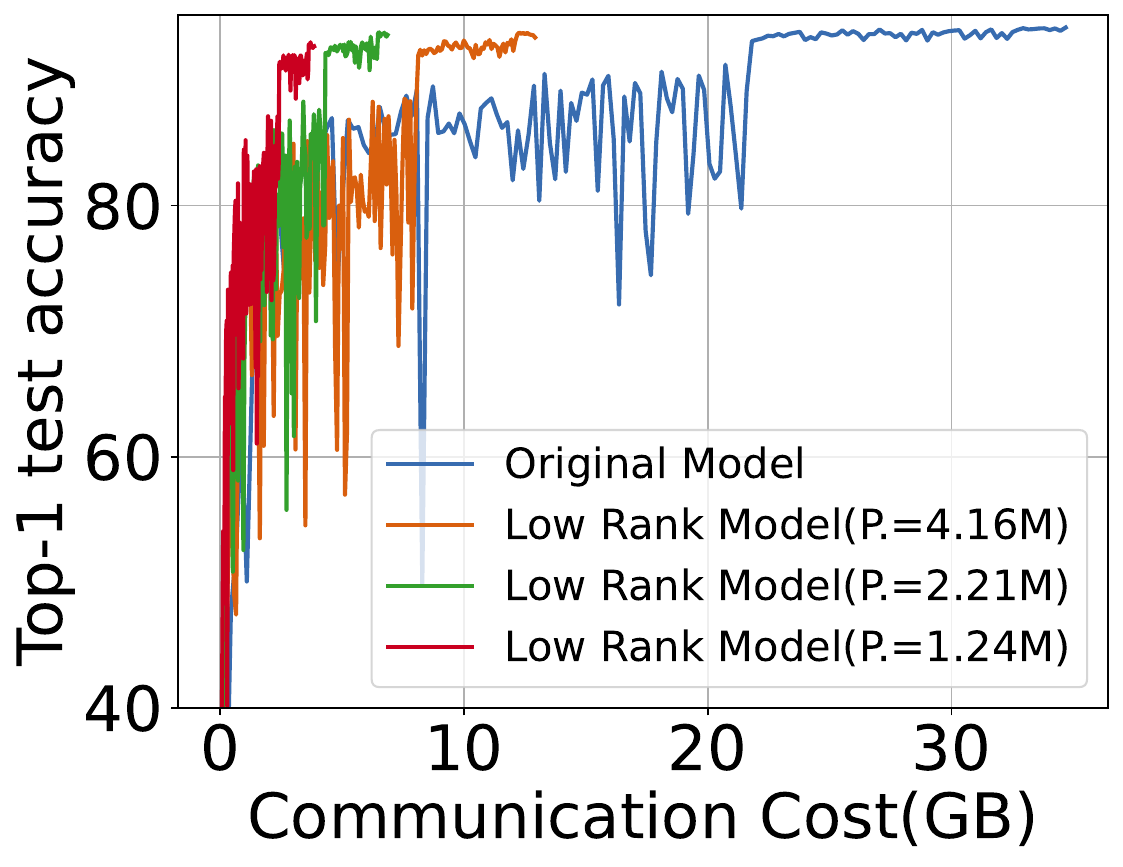} 
\caption{CIFAR-10 IID}
\end{subfigure}
\begin{subfigure}{.245\textwidth}
\includegraphics[width=.99\columnwidth]{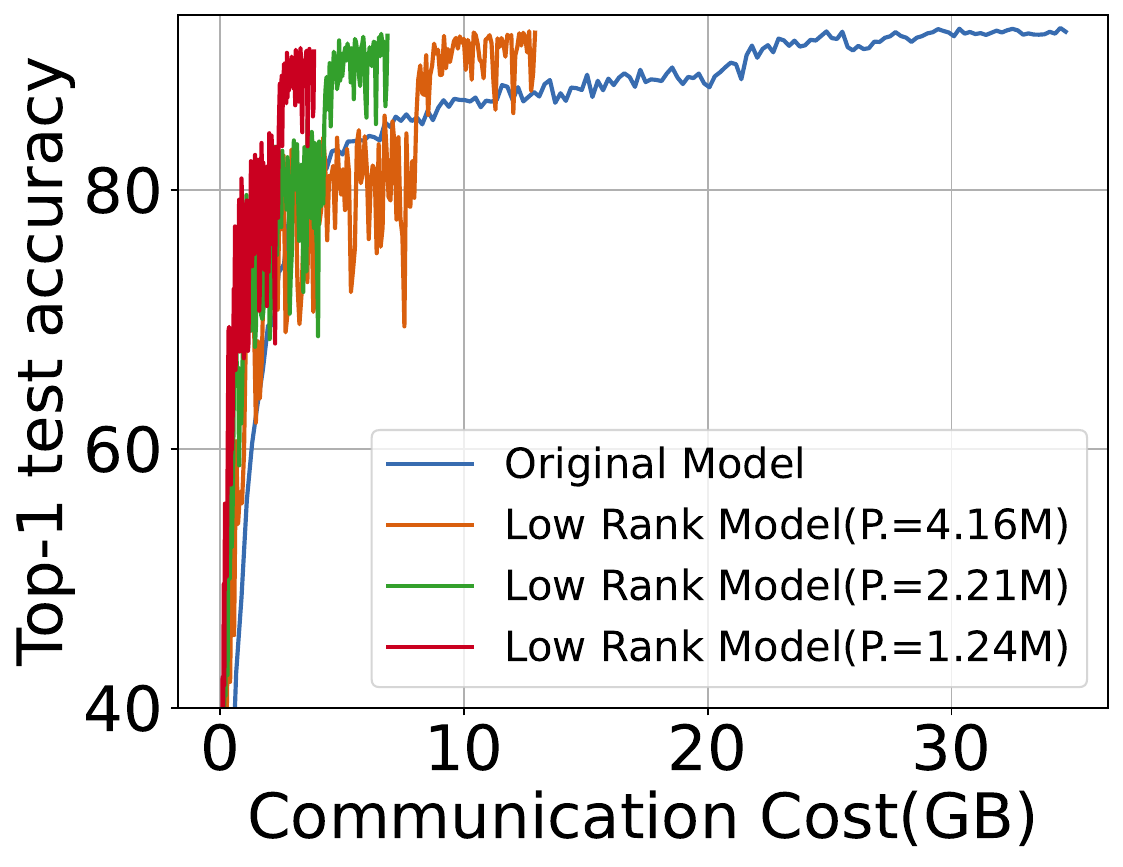}
\caption{CIFAR-10 Non-IID }
\end{subfigure}
\begin{subfigure}{0.245\textwidth}
\includegraphics[width=.99\columnwidth]{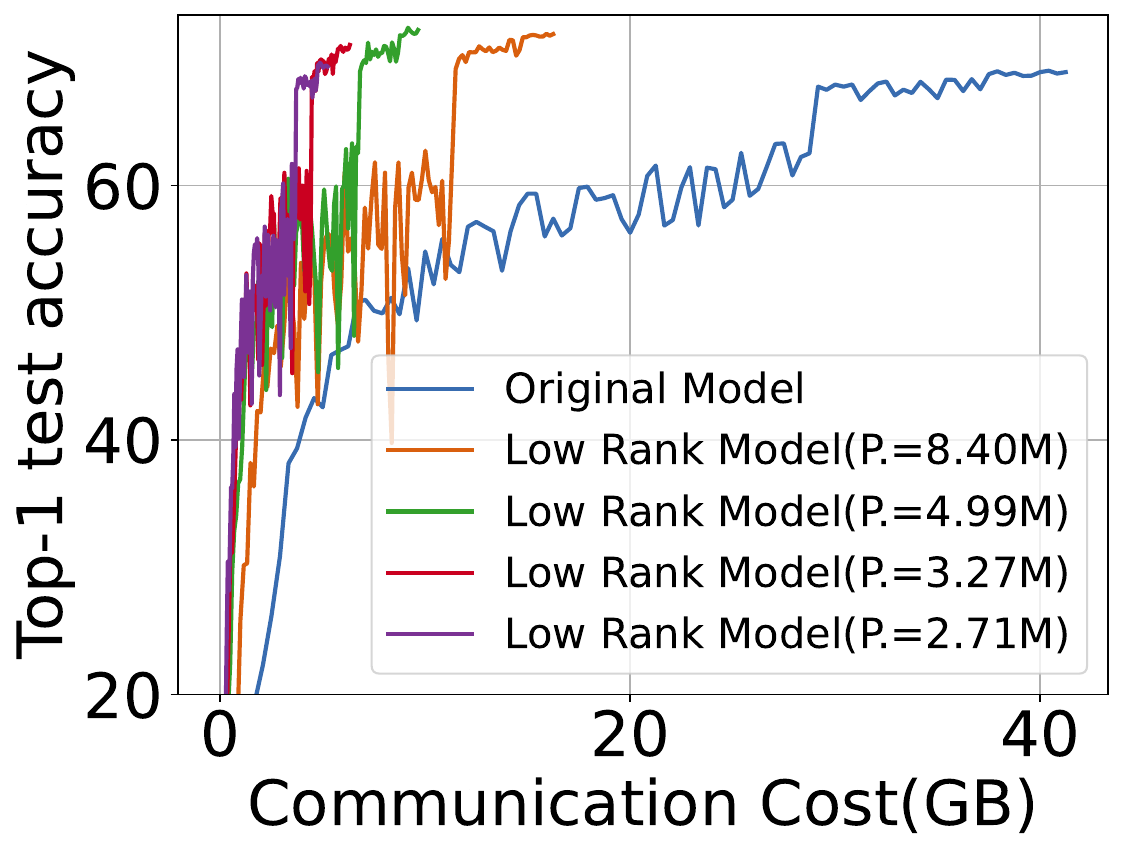} 
\caption{CIFAR-100 IID}
\end{subfigure}
\begin{subfigure}{.245\textwidth}
\includegraphics[width=0.99\columnwidth]{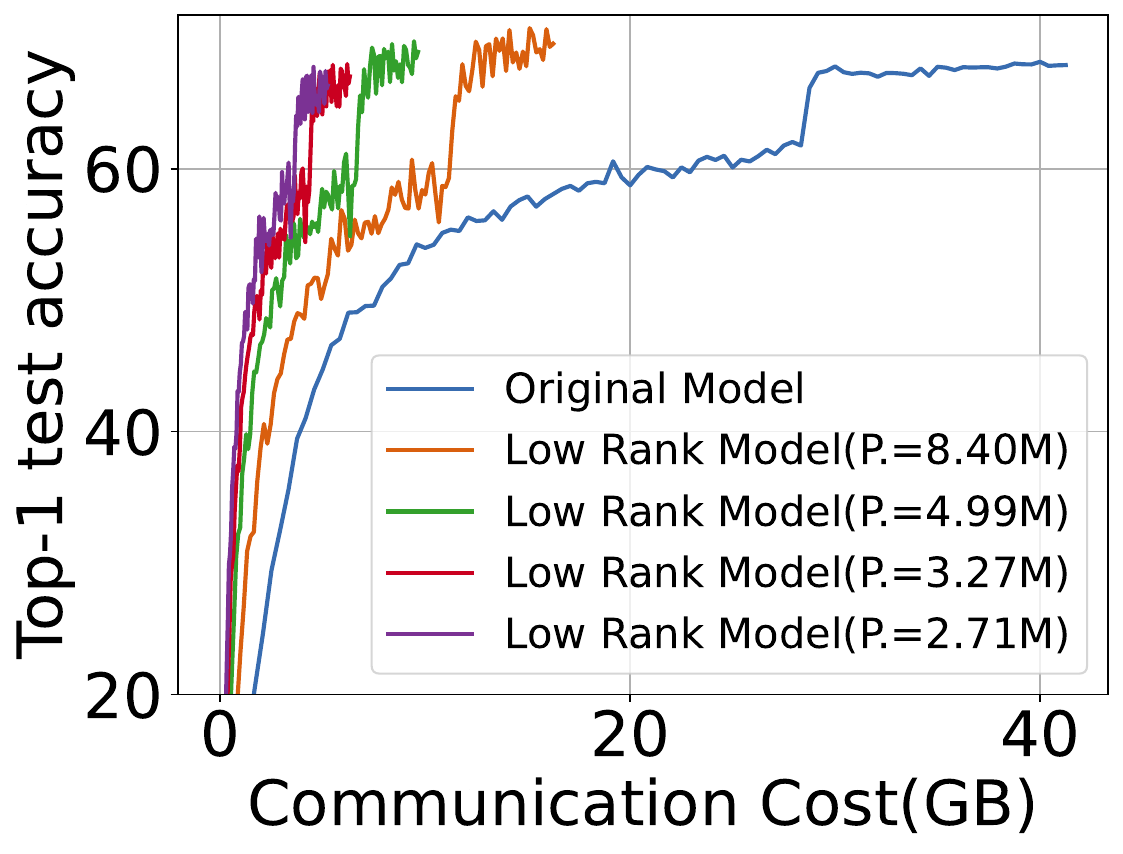}
\caption{CIFAR-100 Non-IID}
\end{subfigure}

\caption{The Top-1 test accuracy v.s. communication cost under homogeneous setting.}
\label{fig:homo_comm}
\end{figure}

\begin{figure}[htb]
\centering

\begin{subfigure}[t]{.48\columnwidth}
\centering
\includegraphics[width=\linewidth]{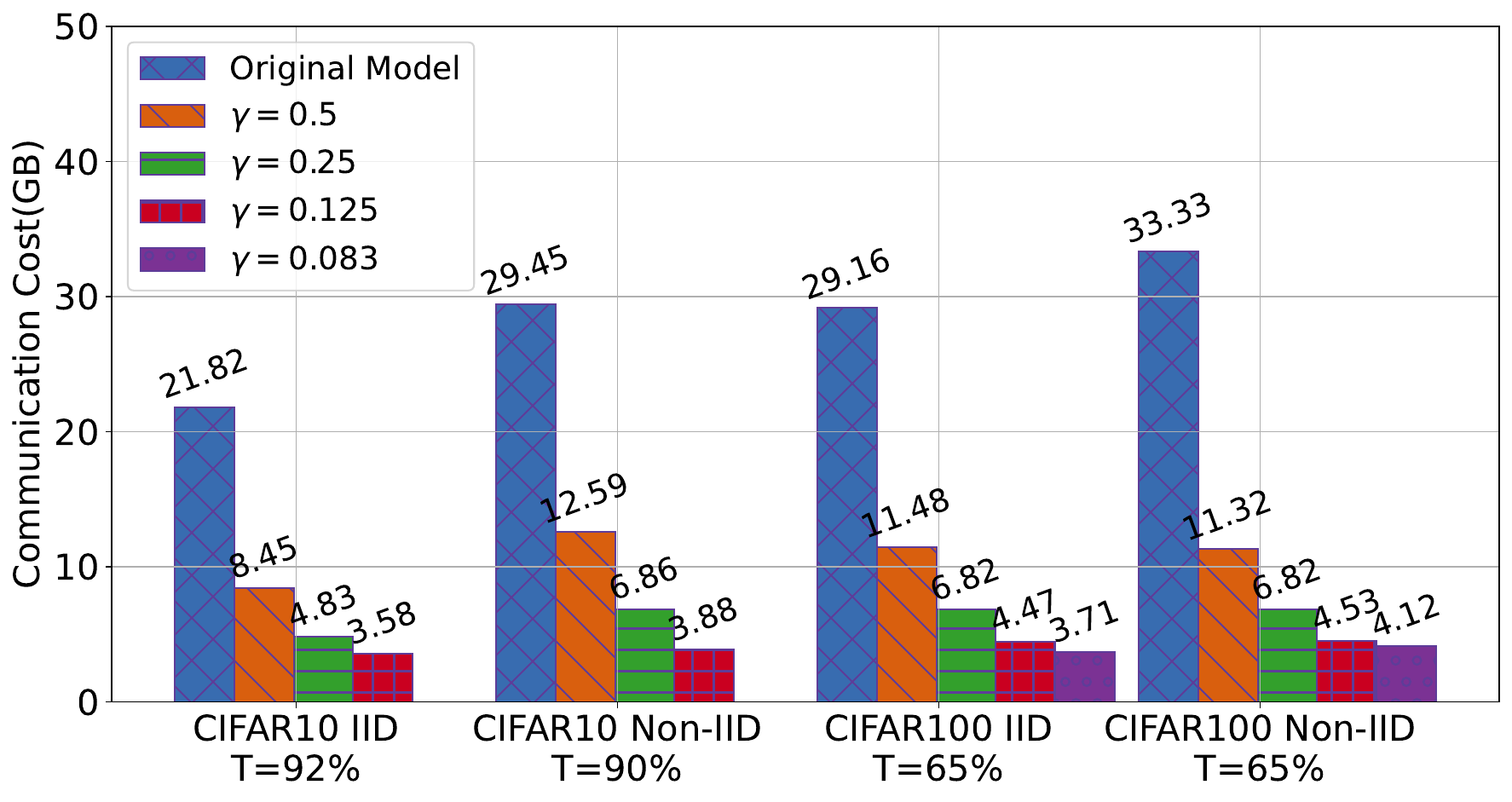}
\caption{Communication cost of \system in FL training.}
\label{fig:target_commcost}
\end{subfigure}
\begin{subfigure}[t]{.48\columnwidth}
\centering
\includegraphics[width=\linewidth]{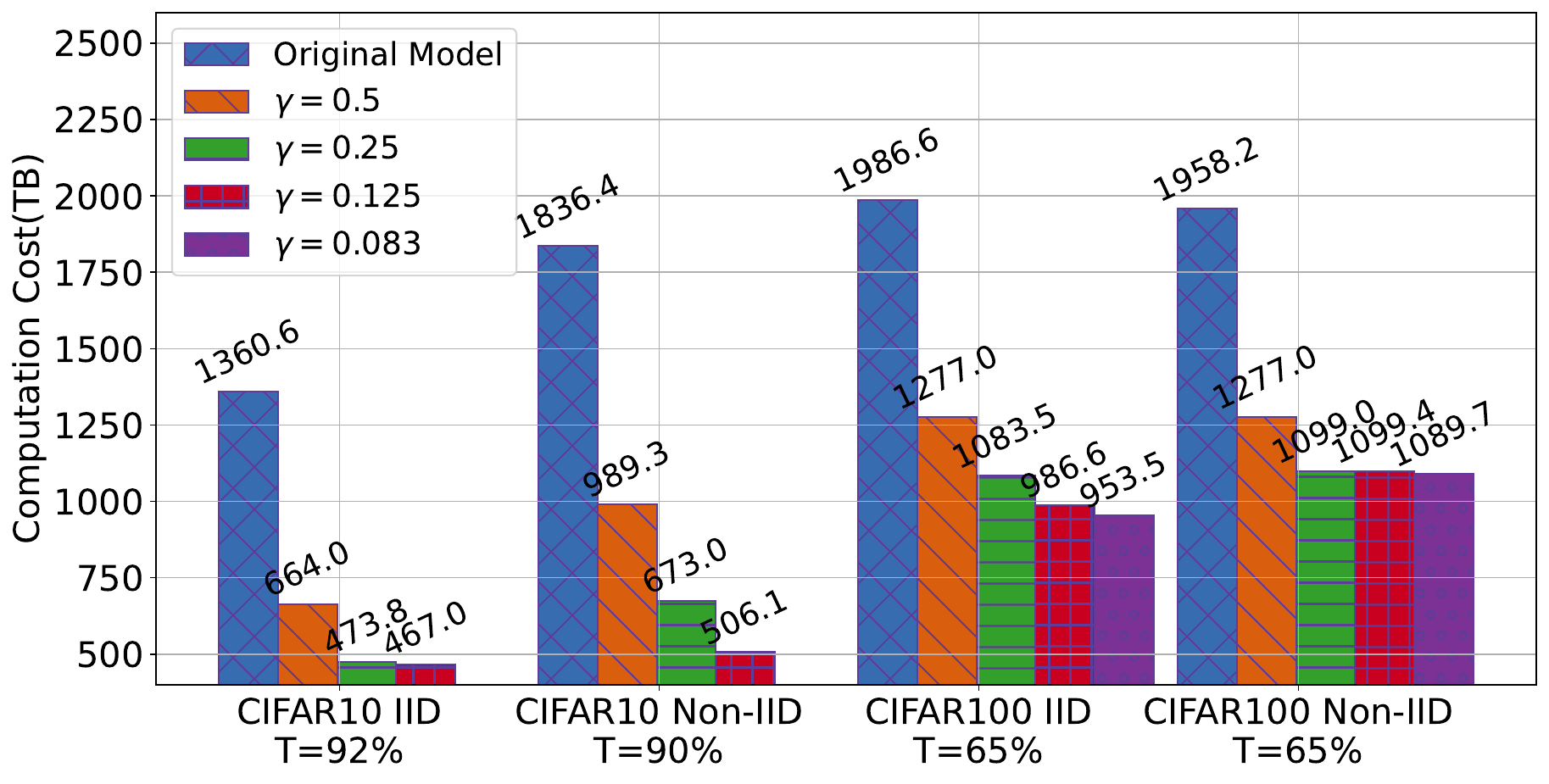}
\caption{Computation cost of \system on each client.}
\label{fig:target_computecost}
\end{subfigure}

\caption{Evaluating different rank ratio $\gamma$ of factorization, in terms of the communication cost and computation cost to reach target Top-1 test accuracy. $T$ denotes the specified target accuracy.}
\label{fig:homo_cost}

\end{figure}

\noindent\textbf{Convergence Speed under Heterogeneous Setting.}
Figure~\ref{fig:hete_iid} shows the convergence speed of \system and the baselines under the fixed heterogeneous setting. Note that, here we consider those models with the smallest size, \eg, 1.37M and 1.24M parameters for baselines and \system, respectively, both on CIFAR-10.
From Figure~\ref{fig:hete_iid}, we find that \system achieves faster convergence than the baselines, with the smallest model of \system consistently outperforming FedAvg, demonstrating the benefits of the factorized training and aggregation strategy. In the early stage of training (training with a larger learning rate), \system holds a more smooth learning curve than FedAvg, which demonstrates that \system improves the robustness of smallest model through aggregation with larger models. 
We also observe that Split-Mix and HeteroFL converges slower than FedAvg,  since both Split-Mix and HeteroFL aim to improve the performance of the server model (the model with the largest size), not the performance of all of the models.

\noindent\textbf{Communication and Computation Cost.}
To evaluate the efficiency of \system, we compare the factorized low-rank models and the original model before factorization, in terms of communication cost and computation cost. The parameters of model and MACs (the number of multiplication-and-addition operations) are key metrics to measure the communication cost and computation overhead. We compute the communication cost as $2\times(\#participants)\times(\#parameters)\times(\#round)$, and the computation overhead is calculated as $(MACs)\times(\#epochs)\times(\#samples)\times(\#round)$.
Figure~\ref{fig:homo_comm} shows the  Top-1 test accuracy \textit{w.r.t.} communication cost under homogeneous setting. 
From this figure, we observe that, compared to original model, \system requires much lower communication cost and achieves comparable Top-1 test accuracy. 
Figure~\ref{fig:target_commcost} shows the communication cost of \system to reach a target accuracy. Compared to the original model, \system reduces the communication cost 6.1$\times$/7.6$\times$ and 7.9$\times$/8.1$\times$, under the CIFAR10 IID/Non-IID and CIFAR100 IID/Non-IID settings.
Figure~\ref{fig:target_computecost} shows that to reach a target accuracy, the original model requires 2.9$\times$/3.6$\times$ and 2.1$\times$/1.8$\times$ more computation cost than \system, under CIFAR10 IID/Non-IID and CIFAR100 IID/Non-IID settings.
The experimental results confirm the effectiveness of local factorized training and that factorized training further improves the performance of client models under the heterogeneous setting.




\section{Related Work}
\label{sec:relwok}
\vspace{-0.5em}

\noindent\textbf{Federated Learning (FL).}
FL distributes a machine learning model to the resource-constrained edges from which the data originates and has emerged as a promising alternative machine learning paradigm~\cite{mcmahan2016communication,mcmahan2016federated,wang2020federated,wang2021field}.
FL enables a multitude of participants to construct a joint model without sharing their private training data~\cite{bonawitz2017practical,liu2021feddg,mcmahan2016communication,mcmahan2016federated}. 
Some recent works focus on model compression or partial network transmission for efficient communication~\cite{albasyoni2020optimal,liang2020think,qiao2021communication,rothchild2020fetchsgd}. However, they do not reduce the computation overhead on edge devices. 
Another line of work aims to offload partial network to the server~\cite{he2020group,wu2021fedadapt} to tackle the computational overhead challenge, but sharing local data information may leak privacy. Our work reduces both the communication cost and training overhead for clients.

\noindent\textbf{Heterogeneous Networks Training.}
\citet{yu2018slimmable} present an approach to training neural networks with different widths and US-Net~\cite{yu2019universally} proposes switchable-BatchNorm to solve the statistics problem in Batch Normalization. Once-for-all~\cite{cai2019once} develops progressive pruning to train the supernet in neural architecture search. DS-Net~\cite{li2021dynamic} proposes to adjust the network according to the input dynamically. The above methods are applicable in model inference. FedMD~\cite{li2019fedmd} and FedDF~\cite{lin2020ensemble} break the knowledge barriers among heterogeneous client models by distillation on proxy data. However, the proxy data is not always available. HeteroFL~\cite{diao2020heterofl} introduces the techniques in~\cite{yu2019universally} into heterogeneous FL and proposes to train networks with different widths in different clients. Split-Mix~\cite{hong2022efficient} proposes to ensemble the small models instead of training large models, while the small models hardly benefit from it. Our proposed method enables the weak devices to benefit from federated learning.

\noindent\textbf{Low-rank Factorization.}
Low-rank factorization is a classical compression method, while it is impractical to factorize the model after training~\cite{lebedev2015speeding,sainath2013low,xue2013restructuring} in FL. Furthermore, directly training the low-rank model from scratch~\cite{konevcny2016federated} results in a performance drop. FedDLR~\cite{qiao2021communication} factorizes the server model and recovers the low-rank model on the clients, while it increases the computation cost. FedPara~\cite{hyeon-woo2022fedpara} proposes a new factorization method, which reduces the communication costs, while does not address the model heterogeneity.
Pufferfish~\cite{wang2021pufferfish} improves the performance of the low-rank model by training a hybrid network and warm-up. \citet{khodak2020initialization} study the gradient descent on factorized layers and propose improving the low-rank models' performance by SI and FD. 
In this paper, we utilize the low-rank factorization to handle the model heterogeneity in federated learning.

\section{Conclusion, Limitations, and Broader Impact}
\label{sec:ckl}
In this paper, we have proposed a novel heterogeneous FL framework \system, which reduces the communication cost and training overhead for the devices. We provide a convergence guarantee for our heterogeneous FL framework. By incorporating low-rank decomposition approaches, \system improves the performance of the server model and client models under various heterogeneous settings. The efficiency of our method can be further improved by combining other efficient FL optimizers.
Comprehensive experiments and analyses verify the effectiveness of our proposed approach when compared with the state-of-the-art baselines. One limitation of \system is that the computation overhead of matrix decomposition is very expensive, though the computation is performed on the server, it is still a burden for the server to perform model factorization at each communication round. A future direction is to use a more efficient model decomposition method for \system. We do not see any negative societal impacts of our work while using our method in practice.
\newpage





{
\small

\bibliographystyle{plainnat}
\bibliography{reference}
}


\newpage
\appendix

\section{More Description of \system}
\subsection{Limitation of FedAvg}
\begin{figure}[hb]
\centering
\begin{minipage}{.5\columnwidth}
    \centering
    \includegraphics[width=\textwidth]{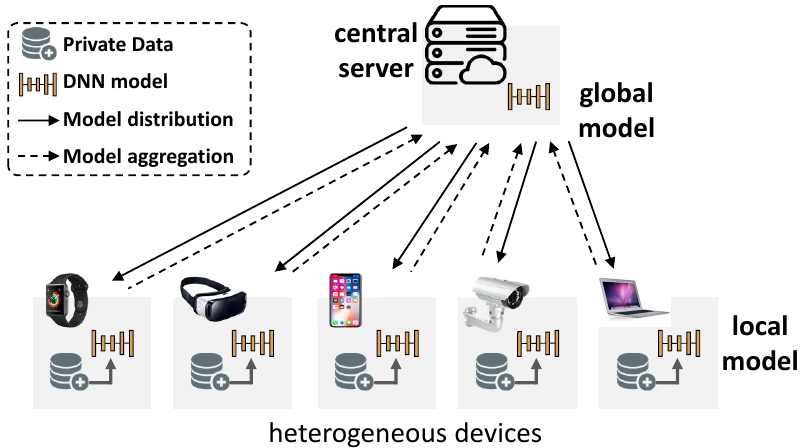}
    \captionof{figure}{The FedAvg framework for heterogeneous computation scenario. }
    \label{fig:modelresource}
\end{minipage}\hfill %
\begin{minipage}{.475\columnwidth}
    \centering
    \includegraphics[width=\linewidth]{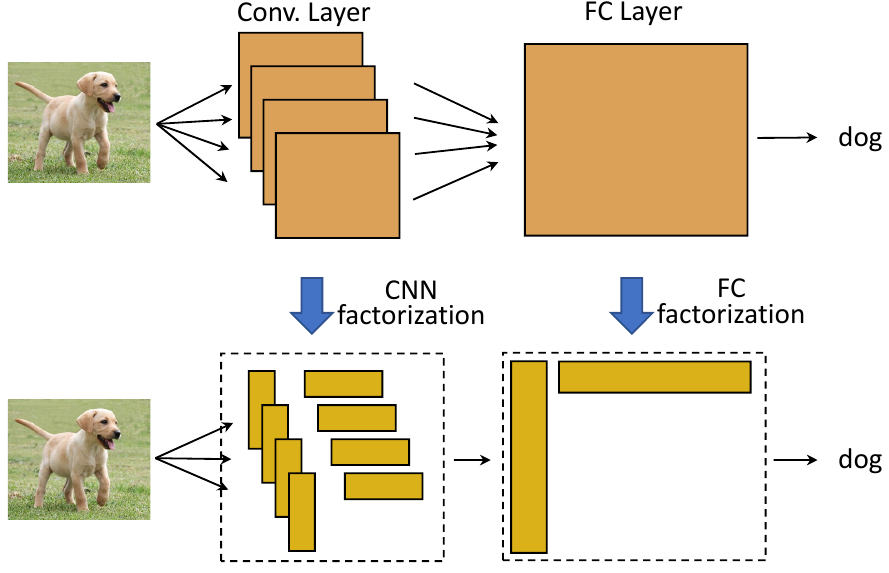}
    \captionof{figure}{FC layer and CNN layer factorization. The number of parameters is significantly reduced.}
    \label{fig:layerfactor}
\end{minipage}
\vspace{-2em}
\end{figure}

From Figure~\ref{fig:modelresource}, we can see that in FedAvg, the local trained models on each device are expected to share the same model architecture, which limits the applied scenarios. 

\subsection{FC Layer Factorization}

Figure~\ref{fig:layerfactor} shows the difference between CNN layer and FC layer factorization.
The forward propagation rule of a fully-connected (FC) layer can be expressed as $\sigma(xW)$ where $x\in \mathbb{R}^{m}, W \in \mathbb{R}^{m\times n}$ and $\sigma(\cdot)$ represents arbitrary non-linear activation function. Then a factorized FC layer can be represented by $\sigma(xUV^\top)$, where $U \in \mathbb{R}^{m\times r}, V^\top \in \mathbb{R}^{r\times n}$, and $r$ denotes the selected rank for the low-rank factorization such that $UV^\top$ is geometrically close to $W$. In the low-rank factorized training, only $U$ and $V^\top$ are held in memory and trained by participating clients. 

\section{Convergence Analysis}
\label{sec:conan}


In this appendix, we provide a proof of Theorem \ref{thm:convergencemain}. We begin by presenting some notation and definitions that will be relevant for our analysis in Section \ref{sec:notation}, then formalize Assumptions \ref{assum:informal1} and \ref{assum:informal2} in Section \ref{sec:assumptions}, and prove our convergence result in Section \ref{sec:convergenceanalysis}.

\subsection{Notation and Definitions} \label{sec:notation}

Throughout this section, we use $w$ to denote variables in the weight space, \ie, $w = \{W^0, \dots, W^L\}$, where $W^\ell$ is the weight matrix at layer $L$ and $h$ to denote variables in the factorized space, \ie, $h = \{W^0, \dots, W^{\rho}, U^{\rho+1}, V^{\rho+1}, \dots, U^{\ell}, V^{\ell}\}$. Occasionally, we convert between these representations. To this end, we define
\begin{equation*}
    w(h) = \{W^0, \dots, W^{\rho}, U^{\rho+1} (V^{\rho+1})^T, \dots, U^{L} (V^L)^T\}.
\end{equation*}
We assume that the quantities $w$ and $h$ have been vectorized for ease of notation, except when working with a specific layer $W^\ell$ or $U^{\ell}$, $V^{\ell}$, in which case we work with the matrix representation. To aid in readability, we largely adopt the notation that vectors are written with lower case letters while matrices are denoted by upper case letters, except in rare cases, where it will be explicitly noted. We use $\|\cdot\|$ to denote the Euclidean norm of a vector and $\|\cdot\|_F$ to denote the Frobenius norm of a matrix. We recall that, for any matrix $A$, $\|\vec(A)\| = \|A\|_F$, where $\vec(A)$ is the vectorization of $A$. In addition, we occasionally make use of the identity
\begin{equation} \label{eq:splitnorm}
    \left\|\left[ v_1^T, \dots, v_n^T \right]^T\right\|^2 = \sum_{i=1}^n \|v_i\|^2
\end{equation}

Now, we recall the definition of $f$ in Eq.~\eqref{eq:fdefmain}:
\begin{equation} \label{eq:fdef}
    \underset{w}{\min} \; f(w) = \frac{1}{P} \sum_{p=1}^P f_p(w),
\end{equation}
where $P$ is the number of participants and each $f_p(w) := \E_{\zeta_p \sim \D_p}[F_p(w;\zeta_p)]$ is a smooth non-convex function where $\D_p$ can vary per participant. Then, we define the factorized loss function, $g$, as
\begin{equation} \label{eq:gdef}
    \underset{h}{\min} \ g(h) = \frac{1}{P} \sum_{p=1}^P g_p(h),
\end{equation}
such that, for any $h$ and $w$ such that $w(h) = w$,
\begin{equation}
    g_p(h) = f_p(w),
\end{equation}
holds for all $p$.

\begin{algorithm}[tb]
\small
\caption{\system: Federated Learning for Heterogeneous Models.}
\label{alg:fedhmsimple}
  \SetKwInOut{KwIn}{Input} \SetKwInOut{KwOut}{Output}
  \KwIn{Dataset $[\D_p]_{p=1}^P$ distributed on $P$ clients, iterations per epoch $E$, total stochastic gradient iterations $T$ such that $T \bmod E = 0$, learning rate $\eta$, rank shrinkage ratios $[r_i]_{i=1}^{\Gamma}$, hyper-parameter $\rho$.}
  \KwOut{Final model $\w_{T+1}$ }
  \nosemic\nonl \textbf{ServerExecute:} \tcp*[h]{server side} \;
  Initialize $L$-layer network $\w_0=\{W_1, \ldots, W_{L}\}$\;
  \For{$t=0,\ldots,T-1$}{
    \If{$t \bmod E = 0$,}{
        \If{$t \neq 0$,}{
            Recover network $w_t$ by $w_t \leftarrow \frac{1}{P} \sum_{p=1}^P w(h_{p,t})$ \;
        }
        \For{$i=1,\ldots,\Gamma$}
        { 
            Factorize the network $w_t$ to a [$\rho$,$L$, $\gamma_i$] hybrid network $\by_t^{\gamma_i}$\;
        }
        Each client $p$, based on capability, chooses some $i\in\{1,\cdots,\Gamma\}$ and sets
        $y_{p,t} \leftarrow \by_t^{\gamma_i}$
            \;
        
        Each client $p$ in parallel draws a random sample $\zeta_{p,t}$ and updates its local solution as
        \begin{equation} \label{eq:hupdate1}
            h_{p,t+1} \leftarrow y_{p,t} - \eta \nabla G(y_{p,t}, \zeta_{p,t})
        \end{equation}
        }
    \Else{
        Each client $p$ in parallel draws a random sample $\zeta_{p,t}$ and updates its local solution
        \begin{equation} \label{eq:hupdate2}
            h_{p,t+1} \leftarrow h_{p,t} - \eta \nabla G(h_{p,t}, \zeta_{p,t}), \; \forall p\;
        \end{equation}
    }
  }
  Recover network $\w_{T}$ by $w_T \leftarrow \frac{1}{P} \sum_{p=1}^P w(h_{p,T})$\;
  \Return $\w_{T}$\;
\end{algorithm}

In order to ease our analysis, we provide a slightly modified version of Algorithm \ref{alg:fedhms} in Algorithm~\ref{alg:fedhmsimple}. Fundamentally, the only difference between these algorithms are:
\begin{itemize}
    \item A modified definition of communication rounds and epoch length, to ease the notation for the iteration counter in the analysis.
    \item Simple averaging as opposed to using soft-max (equivalent to setting $\tau = \infty$ in Eq.~\eqref{eq:fedmc_agg}.)
    \item The use of stochastic gradient steps, as opposed to a generic training algorithm.
\end{itemize}
As such, one can view Algorithm~\ref{alg:fedhmsimple} as a specific instance of Algorithm $\ref{alg:fedhms}$. And we remark that $E$ represents the communication frequency, i.e., clients upload their ``local models" every $E$ steps to the sever to produce a ``global models". We use generic notation $y$ to denote the global models while $h$ to denote the local models (dependence on the iteration and/or clients are made clear via subscripts.)

Now, we describe some basic properties of learning with factorized layers. For all iterations $t$, participants $p$, and layers $\ell \in \{\rho+1,\dots,L\}$, let $U_{p,t}^\ell \in \mathbb{R}^{m \times r}$ and $V_{p,t}^\ell \in \mathbb{R}^{n \times r}$ denote the factors of layer $\ell$ for participant $p$ at time $t$. Then, the layer-wise gradient $\nabla g_{p}^{\ell}(h_{p,t})$ and stochastic gradient $\nabla G_{p}^{\ell}(h_{p,t}; \zeta_{p,t})$ satisfy
\begin{equation} \label{eq:sgUV}
    \nabla g_{p}^{\ell}(h_{p,t}) = \left[\begin{matrix}
    \nabla f_{p}^{\ell}(w_{p,t}) V_{p,t}^{\ell} \\
    \nabla f_{p}^{\ell}(w_{p,t})^T U_{p,t}^{\ell}
    \end{matrix}\right] \ \text{and} \ 
    \nabla G_{p}^{\ell}(h_{p,t}; \zeta_p) = \left[\begin{matrix}
    \nabla F_{p}^{\ell}(w_{p,t};\zeta_{p,t}) V_{p,t}^{\ell} \\
    \nabla F_{p}^{\ell}(w_{p,t};\zeta_{p,t})^T U_{p,t}^{\ell}
    \end{matrix}\right].
\end{equation}
For simplicity of notation, we denote the stochastic gradients at each iteration by $G_{p,t}$, that is
\begin{equation} \label{eq:stochGdef}
    G_{p,t} =
    \begin{cases}
        \nabla G(h_{p,t}, \zeta_{p,t}) & \text{if } t \bmod E \neq 0 \\
        \nabla G(y_{p,t}, \zeta_{p,t}) & \text{if } t \bmod E = 0.
    \end{cases}
\end{equation}
We note that both $\nabla G(h_{p,t}; \zeta_{p,t})$ and $G_{p,t}$ are vectors, even though they are denoted in capital letters.

Throughout this section we assume that we are in the fixed heterogeneous setting, so that the size of $U_{p,t}^{\ell}$ and $V_{p,t}^{\ell}$ are constant at all iterations $t$, for any participant $p$ and layer $\ell \in \{\rho+1,\dots,L\}$. Then, we define $r(p,\ell)$ to be the rank of $U_{p,t}^{\ell}$ and $V_{p,t}^{\ell}$ for all $\ell \in \{\rho+1,\dots,L\}$. Given this, for each layer $\ell \in \{\rho+1,\dots,L\}$, we define the maximum layer rank $\rmaxl$ as
\begin{equation*}
    \rmaxl := \underset{p}{\max}\left(r(p,\ell)\right).
\end{equation*}

The standard approach for analyzing federated learning with full participation (or parallel restarted SGD) relies on analyzing the behavior of the average of the iterates, including at iterations where no averaging occurs. However, direct averaging of the iterates is complicated due to different participants working with models of different sizes. Nevertheless, due to Eq.~\eqref{eq:sgUV}, it follows that one can pad the factors $U_p^{\ell}$ and $V_p^{\ell}$ with matrices of zeros while maintaining the same stochastic gradient iterates. In particular, for a layer $\ell$ with $W^{\ell} \in \mathbb{R}^{m \times n}$, let
\begin{equation} \label{eq:paddediterates}
    \tilde{U}_p^{\ell} = \left[U_p^{\ell}, 0_{m \times (\rmaxl-r(p,\ell))}\right], \ \ \tilde{V}_p^{\ell} = \left[V_p^{\ell}, 0_{n \times (\rmaxl-r(p,\ell))}\right],
\end{equation}
where $0_{i,j}$ is the zero matrix of size $i \times j$. Then,
\begin{align*}
    \left[\begin{matrix}
    \tilde{U}_{p,t+1}^{\ell} \\
    \tilde{V}_{p,t+1}^{\ell} 
    \end{matrix} \right] &= 
    \left[\begin{matrix}
    \tilde{U}_{p,t}^{\ell} \\
    \tilde{V}_{p,t}^{\ell} 
    \end{matrix} \right] - \eta
    \left[\begin{matrix}
    \nabla F_p(w^{\ell}_{p,t}; \zeta_{p,t}) \tilde{V}_{p,t}^{\ell} \\
    \nabla F_p(w^{\ell}_{p,t}; \zeta_{p,t})^T \tilde{U}_{p,t}^{\ell} 
    \end{matrix} \right] \\
    &= 
    \left[\begin{matrix}
    U_{p,t}^{\ell}, 0_{m \times (\rmaxl-r(p,\ell))} \\
    V_{p,t}^{\ell}, 0_{n \times (\rmaxl-r(p,\ell))}
    \end{matrix} \right] - \eta
    \left[\begin{matrix}
    \nabla F_p(w^{\ell}_{p,t}; \zeta_{p,t}) \left[V_{p,t}^{\ell}, 0_{n \times (\rmaxl-r(p,\ell))}\right] \\
    \nabla F_p(w^{\ell}_{p,t}; \zeta_{p,t})^T \left[U_{p,t}^{\ell}, 0_{m \times (\rmaxl-r(p,\ell))}\right]
    \end{matrix} \right] \\
    &= 
    \left[\begin{matrix}
    \left(U_{p,t}^{\ell} - \eta
    \nabla F_p(w^{\ell}_{p,t}; \zeta_{p,t}) V_{p,t+1}^{\ell}\right), 0_{m \times (\rmaxl-r(p,\ell))} \\
    \left(V_{p,t}^{\ell} - \eta
    \nabla F_p(w^{\ell}_{p,t}; \zeta_{p,t})^T U_{p,t+1}^{\ell}\right), 0_{n \times (\rmaxl-r(p,\ell))}
    \end{matrix} \right]
\end{align*}
Therefore, by padding the factors with properly sized matrices of zeros, the behavior of the iterates is unchanged. In addition, by Eq.~\eqref{eq:paddediterates}, it follows that
\begin{equation*}
    \tilde{U}_p^{\ell} \in \mathbb{R}^{m,\rmaxl}, \ \ \tilde{V}_p^{\ell} \in \mathbb{R}^{m,\rmaxl}, \ \forall p, \forall \ell \in\{\rho+1,\dots, L\},
\end{equation*}
so that the matrices $\tilde{U}_p^{\ell}$ and $\tilde{V}_p^{\ell}$ can be directly averaged over the participants $p$. In addition, it should be clear that
\begin{equation*}
    W^{\ell}_p = U^{\ell}_p (V^{\ell}_p)^T = \tilde{U}^{\ell}_p (\tilde{V}^{\ell}_p)^T,
\end{equation*}
so that the inclusion of these zeros has no impact on $W^{\ell}_p$ as well. Throughout the rest of this section, we omit the tildes for ease of presentation and assume that the factorized representation of participant $p$, $h_p$, has been properly padded with zeros to enable direct comparison.

We now define a number of sequences that will be used throughout our analysis. Let the average of the local solutions at time $t$ be defined as
\begin{equation} \label{eq:hbar}
    \bh_t := \frac{1}{P} \sum_{p=1}^P h_{p,t}.
\end{equation}
Recalling the definition of $G_{p,t}$ in Eq.~\eqref{eq:stochGdef}, it follows that, for any iteration where $t \bmod E \neq 0$,
\begin{equation} \label{eq:hbarupdate1}
    \bh_{t+1} = \bh_t - \frac{\eta}{P} \sum_{p=1}^P G_{p,t}.
\end{equation}
In addition, when $t \bmod E = 0$, we define
\begin{equation} \label{eq:ybar}
    \by_t := \frac{1}{P} \sum_{p=1}^P y_{p,t},
\end{equation}
so that
\begin{equation} \label{eq:hbarupdate2}
    \bh_{t+1} = \by_t - \frac{\eta}{P} \sum_{p=1}^P G_{p,t}.
\end{equation}
Finally, we define the sequence $\hat{h}_t$ as
\begin{equation} \label{eq:hath}
    \hat{h}_t := \begin{cases}
        \bh_t & \text{if }t \bmod E \neq 0, \\
        \by_t & \text{if }t \bmod E = 0.
    \end{cases}
\end{equation}

Now, we consider the properties of $\by_t$ and $y_{p,t}$ at any averaging step. For any $\ell \in \{\rho+1,\dots,L\}$, let $U_t^{\ell} \Sigma_t^{\ell} (V_t^{\ell})^T$ be the singular value decomposition of $W_t^{\ell}$, where $U_t^{\ell}$ and $V_t^{\ell}$ are orthogonal matrices and $\Sigma_t^{\ell}$ is the diagonal matrix of the singular values of $W_t^{\ell}$. Then, defining $\Sigma_{r(p,\ell),t}^{\ell} = \diag(\sigma_1,\dots,\sigma_{r(p,\ell)},0,\dots, 0)$, for any $y_{p,t}^{\ell}$,
 \begin{equation} \label{eq:ypt}
    y_{p,t}^{\ell} = \vec\left(\left[ \begin{matrix} U_t^{\ell} \sqrt{\Sigma_{r(p,\ell),t}^{\ell}} \\ V_{t}^{\ell} \sqrt{\Sigma_{r(p,\ell),t}^{\ell}} \end{matrix} \right]\right) \ \text{ and } \ \by_t^{\ell} = \vec\left(\left[ \begin{matrix} \frac{1}{P} \sum_{p=1}^P U_t^{\ell} \sqrt{\Sigma_{r(p,\ell),t}^{\ell}} \\ \frac{1}{P} \sum_{p=1}^P V_{t}^{\ell} \sqrt{\Sigma_{r(p,\ell),t}^{\ell}} \end{matrix}\right]\right).
\end{equation}
For completeness, we define
\begin{equation*}
    \bh_0 := \vec\left(\left\{W^0, \dots, W^{\rho}, U_0^{\rho+1} \sqrt{\Sigma_0^{\rho+1}}, V_0^{\rho+1} \sqrt{\Sigma_0^{\rho+1}}, \dots, U_0^L \sqrt{\Sigma_0^L}, V_0^L \sqrt{\Sigma_0^L}\right\}\right),
\end{equation*} so that $g(\bh_0) = f(w_0)$.

Finally, we use $\E_{p,t}[\cdot]$ to denote an expected value of the random variable $\zeta_{p,t}$ conditioned on the full history of the randomness up to the start of iteration $t$, which we denote by $\zeta_{[t-1]} := [\zeta_{p,\hat{t}}]_{p \in \{1,\dots,P\}, \hat{t} \in \{1,\dots,t-1\}}$. Specifically, we let
\begin{equation*}
    \E_{p,t}[\cdot] = \E_{\zeta_{p,t} \sim \D_p}[\cdot | \zeta_{[t-1]}].
\end{equation*}

\subsection{Assumptions}
\label{sec:assumptions}

Using the notation defined in the previous section, we are prepared to state our assumptions. Our first assumption is frequently used in the non-convex federated learning literature.

\begin{assumption} \label{assum:fsmooth}
    Each function $f_p(w)$ is continuously differentiable, the gradients of $f_p(w)$ are Lipschitz with modulus $L_{\nabla f}$, and the function $f(w)$ is bounded below by $\flow \in \mathbb{R}$. For each participant $p$ and iteration $t$, the stochastic gradients satisfy
    \begin{equation*}
        \E_{p,t}[\nabla F_p(w_{p,t}; \zeta_{p,t})] = \nabla f_p(w_{p,t}).
    \end{equation*}
    In addition, there exist constants $M_f > 0$ and $\kappa_f > 0$ such that
    \begin{align*}
        &\E_{\zeta_p \sim \D_p}[ \|\nabla F_p(w; \zeta_p) - \nabla f_p(w)\|^2]  \leq M_f, \ \forall w, \forall p \\
        &\E_{\zeta_p \sim \D_p} [\|\nabla F_p(w; \zeta_p)\|^2] \leq \kappa_f, \ \forall w, \forall p.
    \end{align*}
\end{assumption}

We note here that while the assumption on bounded second moments is strong, it is a common assumption in the literature for non-convex parallel restarted SGD \cite{yu2019parallel}.

Recalling Eq.~\eqref{eq:sgUV}, it follows that under Assumption \ref{assum:fsmooth},
\begin{equation*}
    \E_{p,t}[\nabla G_p(h_{p,t}; \zeta_{p,t})] = \nabla g_p(h_{p,t}).
\end{equation*}
However, Assumption \ref{assum:fsmooth} is insufficient to imply that the gradients of $g_p$ are Lipschitz continuous gradients nor that the stochastic gradients have bounded variance, and bounded second moments. This motivates the following assumption.

\begin{assumption} \label{assum:factorsbounded}
    There exists a constant $\kappa_{UV} > 0$ such that, at every iteration $t$ of Algorithm \ref{alg:fedhmsimple}, 
    \begin{equation*}
        \|U_{p,t}^{\ell}\|_F \leq \kappa_{UV}, \quad \|V_{p,t}^{\ell}\|_F \leq \kappa_{UV}, \; \forall p, \forall \ell \in \{\rho+1,\dots,L\},
    \end{equation*}
    where $\|\cdot\|_F$ denotes the Frobenius norm.
\end{assumption}

Under Assumptions \ref{assum:fsmooth} and \ref{assum:factorsbounded}, it follows that the gradients of $g_p(h)$ are Lipschitz continuous for some modulus $\LUV > 0$ and there exist constants $M_g > 0$ and $\kappa_g > 0$ such that,
\begin{align*}
    &\E_{\zeta_{p} \sim \D_p} [\|\nabla G_p(h; \zeta_p) - \nabla g_p(h)\|^2] \leq M_g, \ \forall h, \forall p, \\
    &\E_{\zeta_{p} \sim \D_p} \|\nabla G_p(h; \zeta_p)\|^2 \leq \kappa_g, \ \forall h, \forall p.
\end{align*}

Most of our results will be proven under the previous two assumptions. However, for the final result, we require an additional assumption that allows us to bound the difference between the server model and the low-rank models at each aggregation step.

\begin{assumption} \label{assum:boundedprojection}
    There exist constants $\kappa_{\sigma_1}$ and $\kappa_{\sigma_2}$ such that, at every iteration $t$ such that $t \bmod E = 0$,
    \begin{equation*}
        \E\left[\left\|\left(\frac{1}{P} \sum_{p=1}^P \sqrt{\Sigma_{r(p,\ell),t}^{\ell}}\right)\left(\frac{1}{P} \sum_{p=1}^P \sqrt{\Sigma_{r(p,\ell),t}^{\ell}}\right) - \Sigma_t^{\ell}\right\|^2_F\right] \leq \kappa_{\sigma_1}
    \end{equation*}
    and
    \begin{equation*}
       \E\left[\left\|\frac{1}{P} \sum_{i=1}^P \sqrt{\Sigma_{r(i,\ell),t}^{\ell}}  - \sqrt{\Sigma_{r(p,\ell),t}^{\ell}}\right\|^2_F\right] \leq \kappa_{\sigma_2}
    \end{equation*}
    In addition, the function $f$ is Lipschitz continuous with modulus $L_f$.
\end{assumption}
Recalling Eq.~\eqref{eq:ypt}, the first inequality in this assumption assumes that the expected norm difference between $w(\by_t)^{\ell}$ and $w_t^{\ell}$ is bounded while the second assumes that, in expectation, the norm difference of the factors of $y_{p,t}^{\ell}$ and $\by_{t}^{\ell}$ are bounded as well. Essentially, these two inequalities state that the expected difference between the ``true" model and the average of the low-rank models is bounded, as well as the difference between the average of the low-rank models and any participant's low-rank model. We remark without such an assumption, there is no reason to expect that Algorithms \ref{alg:fedhms} and \ref{alg:fedhmsimple} would produce a reasonable model, as arbitrary error may be added during the network factorization step.

We note that the final assumption, relating to the Lipschitz continuity of $f$ is essentially implied by Assumption \ref{assum:fsmooth}, as the bounded second moments imply that the gradient of $f$ is bounded as well. A well-known consequence of boundedness of the gradient is Lipschitz continuity of the function. Therefore, while, in essence, we could use $\kappa_g$ as a Lipschitz constant for $f$, we differentiate these two quantities to make their contribution to the final bound more clear.

\subsection{Convergence Proofs} \label{sec:convergenceanalysis}

In this section, we prove Theorem \ref{thm:convergencemain}. We begin with a number of preliminary lemmas and note that the structure of our analysis is heavily based on the analysis of parallel restarted SGD found in \cite{yu2019parallel}. First, we derive a preliminary result about the drift between individual models $h_{p,t}$ and the model average $\bh_t$.

\begin{lemma} \label{lem:drift}
    Let Assumptions \ref{assum:fsmooth} and \ref{assum:factorsbounded} hold. Then, for any participant $p$ and $t > 0$,
    \begin{equation*}
        \E[\|\bh_t - h_{p,t}\|^2] \leq 6 \eta^2 E^2 \kappa_g + 3 \E[\|\by_{t_0} - y_{p,t_0}\|^2],
    \end{equation*}
    where $t_0$ is the largest $t_0 < t$ such that $t_0 \bmod E = 0$.
\end{lemma}

\begin{proof}
    By the updates of Eq.~\eqref{eq:hupdate1}, Eq.~\eqref{eq:hupdate2}, and the definition of the stochastic gradients Eq.~\eqref{eq:sgUV}, we have
    \begin{equation*}
        h_{p,t} = y_{p,t_0} - \eta \sum_{\hat{t} = t_0}^{t-1} G_{p,\hat{t}}.
    \end{equation*}
    Similarly, by Eq.~\eqref{eq:hbarupdate1} and Eq.~\eqref{eq:hbarupdate2},
    \begin{equation*}
        \bh_t = \by_{t_0} - \eta \sum_{\hat{t} = t_0}^{t-1} \frac{1}{P} \sum_{i=1}^P G_{i,\hat{t}}.
    \end{equation*}
    Therefore, it follows that
    \begin{align*}
        &\E[\|\bh_t-h_{p,t}\|^2]
        = \E\left[\left\|\eta \sum_{\hat{t} = t_0}^{t-1} \frac{1}{P} \sum_{i=1}^P G_{i,\hat{t}} - \eta \sum_{\hat{t} = t_0}^{t-1} G_{p,\hat{t}} + \by_{t_0} - y_{p,t_0}\right\|^2\right] \\
        &\leq 3\eta^2 \E\left[\left\|\sum_{\hat{t} = t_0}^{t-1} \frac{1}{P} \sum_{i=1}^P G_{i,\hat{t}}\right\|^2 + \left\|\sum_{\hat{t} = t_0}^{t-1} G_{p,\hat{t}}\right\|^2\right] + 3\E[\left\|\by_{t_0} - y_{p,t_0}\right\|^2] \\
        &\leq 3\eta^2(t-t_0) \E\left[\sum_{\hat{t} = t_0}^{t-1} \left\|\frac{1}{P} \sum_{i=1}^P G_{i,\hat{t}}\right\|^2 + \sum_{\hat{t} = t_0}^{t-1} \left\| G_{p,\hat{t}}\right\|^2\right] + 3\E[\left\|\by_{t_0} - y_{p,t_0}\right\|^2] \\
        &\leq 3\eta^2(t-t_0) \left(\sum_{\hat{t} = t_0}^{t-1} \frac{1}{P} \sum_{i=1}^P \E\left[\left\| G_{i,\hat{t}}\right\|^2\right] + \sum_{\hat{t} = t_0}^{t-1} \E\left[\left\| G_{p,\hat{t}}\right\|^2\right]\right) + 3\E[\left\|\by_{t_0} - y_{p,t_0}\right\|^2] \\
        &\leq 6\eta^2 E^2 \kappa_g + 3\E[\left\|\by_{t_0} - y_{p,t_0}\right\|^2],
    \end{align*}
    where the first three inequalities follow by $\|\sum_{i=1}^n z_i\|^2 \leq n \sum_{i=1}^n \|z_i\|^2$ for any vectors $z_i$ ($n = 3$ for the first inequality, $n = t-t_0$ for the second inequality, and $n = P$ for the third inequality) and the final inequality follows by Assumptions \ref{assum:fsmooth} and \ref{assum:factorsbounded} and $t-t_0 \leq E$.
\end{proof}

Next, we a preliminary lemmas which bounds the expected difference between the average sequence $\bh$ at subsequent iterates.

\begin{lemma} \label{lem:hbardiff}
    Let Assumptions \ref{assum:fsmooth} and \ref{assum:factorsbounded}. Then, when $t \bmod E \neq 0$,
    \begin{equation*}
        \E[\|\bh_{t+1} - \bh_t\|^2] \leq \frac{\eta^2 M_g}{P} + \eta^2 \E\left[\left\|\frac{1}{P} \sum_{p=1}^P \nabla g_p(h_{p,t}) \right\|^2 \right].
    \end{equation*}
    Otherwise, when $t \bmod E = 0$,
    \begin{equation*}
        \E[\|\bh_{t+1} - \by_t\|^2] \leq \frac{\eta^2 M_g}{P} + \eta^2 \E\left[\left\|\frac{1}{P} \sum_{p=1}^P \nabla g_p(y_{p,t}) \right\|^2 \right].
    \end{equation*}
\end{lemma}

\begin{proof}
    To start, let $t \bmod E \neq 0$ so that $\hat{h}_t = \bh_t$ due to Eq.~\eqref{eq:hath}.  Then, by Eq.~\eqref{eq:hbarupdate1}, it follows that
    \begin{equation} \label{eq:hbardiffbound1}
        \E[\|\bh_{t+1} - \bh_t\|^2] = \eta^2 \E\left[\left\|\frac{1}{P}\sum_{p=1}^P G_{p,t}\right\|^2\right].
    \end{equation}
    Next, by the linearity of the expectation, we have,
    \begin{align}
        &\E\left[\left\|\frac{1}{P} \sum_{p=1}^P G_{p,t} \right\|^2 \right]
        = \E\left[\left\|\frac{1}{P} \left(\sum_{p=1}^P  G_{p,t} - \nabla g_p(h_{p,t})\right) + \frac{1}{P} \sum_{p=1}^P \nabla g_p(h_{p,t}) \right\|^2 \right] \nonumber \\
        &= \E\left[\left\|\frac{1}{P} \left(\sum_{p=1}^P  G_{p,t} - \nabla g_p(h_{p,t})\right)\right\|^2\right] + \E\left[\left\|\frac{1}{P} \sum_{p=1}^P \nabla g_p(h_{p,t}) \right\|^2 \right] \nonumber \\
        &\quad + \frac{2}{P^2} \E\left[\left(\sum_{p=1}^P G_{p,t} - \nabla g_p(h_{p,t})\right)^T \left( \sum_{p=1}^P \nabla g_p(h_{p,t})\right)\right]. \label{eq:hbardiffbound2}
    \end{align}
    Recalling that $\E_{p,t}[G_{p,t}] = \E_{p,t}[\nabla G(h_{p,t};\zeta_{p,t})] = \nabla g_p(h_{p,t})$, by the law of iterated expectation,
    \begin{align}
        &\E\left[\left(\sum_{p=1}^P G_{p,t} - \nabla g_p(h_{p,t})\right)^T \left(\sum_{p=1}^P \nabla g_p(h_{p,t})\right)\right] \nonumber \\
        &=\E\left[\E\left[\left(\sum_{p=1}^P G_{p,t} - \nabla g_p(h_{p,t})\right)^T \left(\sum_{p=1}^P \nabla g_p(h_{p,t})\right)\Bigg| \zeta_{[t-1]}\right]\right] \nonumber \\
        &=\E\left[\left(\sum_{p=1}^P \E_{p,t}\left[G_{p,t}\right] - \nabla g_p(h_{p,t})\right)^T \left(\sum_{p=1}^P \nabla g_p(h_{p,t})\right)\right] \nonumber \\
        &=\E\left[\left(\sum_{p=1}^P \nabla g_p(h_{p,t}) - \nabla g_p(h_{p,t})\right)^T \left(\sum_{p=1}^P \nabla g_p(h_{p,t})\right)\right] = 0, \label{eq:hbardiffbound3}
    \end{align}
    where the second equality follows by the definition of $\E_{p,t}[\cdot]$ and the fact that $h_{p,t}$ is deterministic when conditioned on $\zeta_{[t-1]}$.
    
    Now, once again using the law of iterated expectation, the fact that $h_{p,t}$ is deterministic when conditioned on $\zeta_{[t-1]}$ and the independence of $G_{p,t}$ across participants,
    \begin{align*}
        &\E\Bigg[\Bigg\|\frac{1}{P} \Bigg(\sum_{p=1}^P  G_{p,t} - \nabla g_p(h_{p,t})\Bigg)\Bigg\|^2\Bigg] \\
        &= \frac{1}{P^2} \E\left[\left(\sum_{p=1}^P  G_{p,t} - \nabla g_p(h_{p,t})\right)^T \left(\sum_{p=1}^P G_{p,t} - \nabla g_p(h_{p,t})\right) \right] \\
        &= \frac{1}{P^2} \E\Bigg[\sum_{p=1}^P  \Bigg\|G_{p,t} - \nabla g_p(h_{p,t})\Bigg\|^2 + 2  \sum_{p=1}^P \Bigg(\Bigg( G_{p,t} - \nabla g_p(h_{p,t})\Bigg)^T \Bigg(\sum_{\hat{p} \neq p}  G_{\hat{p},t} - \nabla g_{\hat{p}}(h_{\hat{p},t})\Bigg)\Bigg) \Bigg] \\
        &= \frac{1}{P^2} \sum_{p=1}^P  \E\left[\left\|G_{p,t} - \nabla g_p(h_{p,t})\right\|^2\right] \\
        &\quad+ \frac{2}{P^2}\E\left[\sum_{p=1}^P \left(\left(\E_{p,t}[G_{p,t}] - \nabla g_p(h_{p,t})\right)^T \left(\sum_{\hat{p} \neq p}  \E_{\hat{p},t}[G_{\hat{p},t}] - \nabla g_{\hat{p}}(h_{\hat{p},t})\right)\right) \right] \\
        &= \frac{1}{P^2} \sum_{p=1}^P  \E\left[\left\|G_{p,t} - \nabla g_p(h_{p,t})\right\|^2\right].
    \end{align*}
    
    Combining this with Eq.~\eqref{eq:hbardiffbound1}, Eq.~\eqref{eq:hbardiffbound2}, and Eq.~\eqref{eq:hbardiffbound3} we have
    \begin{align*}
        \E[\|\bh_{t+1} - \bh_t\|^2] &= \eta^2 \E\left[\left\|\frac{1}{P} \left(\sum_{p=1}^P  G_{p,t} - \nabla g_p(h_{p,t})\right)\right\|^2\right] + \eta^2 \E\left[\left\|\frac{1}{P} \sum_{p=1}^P \nabla g_p(h_{p,t}) \right\|^2 \right] \\
        &= \frac{\eta^2}{P^2} \sum_{p=1}^P \E\left[\left\|G_{p,t} - \nabla g_p(h_{p,t})\right\|^2\right] + \eta^2 \E\left[\left\|\frac{1}{P} \sum_{p=1}^P \nabla g_p(h_{p,t}) \right\|^2 \right] \\
        &\leq \frac{\eta^2 M_g}{P} + \eta^2 \E\left[\left\|\frac{1}{P} \sum_{p=1}^P \nabla g_p(h_{p,t}) \right\|^2 \right],
    \end{align*}
    where $M_g$ is the bound on the variance implied by Assumptions \ref{assum:fsmooth} and \ref{assum:factorsbounded}. This proves the result for $t \bmod E \neq 0$.
    
    On the other hand, when $t \bmod E = 0$, by Eq.~\eqref{eq:hbarupdate2},
    \begin{equation*}
        \E[\|\bh_{t+1} - \by_t\|^2] = \eta^2 \E\left[\left\|\frac{1}{P}\sum_{p=1}^P G_{p,t}\right\|^2\right],
    \end{equation*}
    and therefore the result follows by exactly the same argument as above with the replacement of $h_{p,t}$ by $y_{p,t}$.
\end{proof}

Next, we provide a bound on the expectation of the inner product of the average gradient and the difference of the averaged iterates.
\begin{lemma} \label{lem:hbarinner}
    Let Assumptions \ref{assum:fsmooth} and \ref{assum:factorsbounded}. Then, for any iteration $t$ such that $t \bmod E \neq 0$,
    \begin{align*}
        \E[\nabla g(\bh_t)^T(\bh_{t+1} - \bh_t)] &\leq -\frac{\eta}{2} \E[\|\nabla g(\bh_t)\|^2] - \frac{\eta}{2} \E\left[ \left\|\frac{1}{P} \sum_{p=1}^P \nabla g_p(h_{p,t})\right\|^2 \right] \\
        &\quad+ 3 \eta^3 \LUV^2 \kappa_g E^2 + \frac{3\eta \LUV^2}{2P} \sum_{p=1}^P \E[\|\by_{t_0} - y_{p,t_0}\|^2],
    \end{align*}
    where $t_0$ is the largest $t_0 < t$ such that $t_0 \bmod E = 0$. In addition, when $t \bmod E = 0$,
    \begin{align*}                  E[\nabla g(\by_t)^T(\bh_{t+1} - \by_t)] &\leq -\frac{\eta}{2} \E[\|\nabla g(\by_t)\|^2] - \frac{\eta}{2} \E\left[ \left\|\frac{1}{P} \sum_{p=1}^P \nabla g_p(y_{p,t})\right\|^2 \right] \\
     &\quad+ \frac{\eta \LUV^2}{2P} \sum_{p=1}^P \E[\|\by_t - y_{p,t}\|^2].
    \end{align*}
\end{lemma}

\begin{proof}
    To start, for $t \bmod E \neq 0$, by Eq.~\eqref{eq:hbarupdate1},
    \begin{equation*}
        \E[\nabla g(\bh_t)^T(\bh_{t+1} - \bh_t)] = - \eta \E\left[\nabla g(\bh_t)^T\left(\frac{1}{P} \sum_{p=1}^P G_{p,t}\right)\right].
    \end{equation*}
    Using the law of iterated expectations and the fact that the stochastic gradients are unbiased estimators of the true gradients,
    \begin{align*}
        \E\left[\nabla g(\bh_t)^T\left(\frac{1}{P} \sum_{p=1}^P G_{p,t} \right)\right] &= \E\left[\E\left[\nabla g(\bh_t)^T\left(\frac{1}{P} \sum_{p=1}^P G_{p,t}\right) \Bigg| \zeta_{t-1}\right]\right] \\
        &= \E\left[\nabla g(\bh_t)^T\left(\frac{1}{P} \sum_{p=1}^P \E_{p,t}[G_{p,t}]\right)\right] \\
        &= \E\left[\nabla g(\bh_t)^T\left(\frac{1}{P} \sum_{p=1}^P \nabla g_p(h_{p,t})\right)\right].
    \end{align*}
    Thus, it follows that
    \begin{align*}
        -\eta \E[\nabla g(\bh_t)^T(\bh_{t+1} - \bh_t)] &= -\eta \E\left[\nabla g(\bh_t)^T\left(\frac{1}{P} \sum_{p=1}^P \nabla g_p(h_{p,t})\right)\right] \\
        &= -\frac{\eta}{2} \E[\|\nabla g(\bh_t)\|^2] - \frac{\eta}{2} \E\left[ \left\|\frac{1}{P} \sum_{p=1}^P \nabla g_p(h_{p,t})\right\|^2 \right] \\
        &\quad+ \frac{\eta}{2} \E\left[\left\|\nabla g(\bh_t) - \frac{1}{P} \sum_{p=1}^P \nabla g_p(h_{p,t}) \right\|^2 \right],
    \end{align*}
    where the second equality follows from $a^T b = (\|a\|^2 + \|b\|^2 - \|a-b\|^2)/2$ for any vectors $a$ and $b$ of the same length. Working with the final term,
    \begin{align*}
        \E\left[\left\|\nabla g(\bh_t) - \frac{1}{P} \sum_{p=1}^P \nabla g_p(h_{p,t}) \right\|^2 \right] &= \E\left[\left\|\frac{1}{P} \sum_{p=1}^P \nabla g_p(\bh_t) - \frac{1}{P} \sum_{p=1}^P \nabla g_p(h_{p,t}) \right\|^2 \right] \\
        &= \frac{1}{P^2} \E\left[\left\|\sum_{p=1}^P \left(\nabla g_p(\bh_t) - \nabla g_p(h_{p,t})\right) \right\|^2 \right] \\
        &\leq \frac{1}{P} \sum_{p=1}^P \E\left[\left\|\nabla g_p(\bh_t) - \nabla g_p(h_{p,t}) \right\|^2 \right] \\
        &\leq \frac{\LUV^2}{P} \sum_{p=1}^P \E\left[\left\|\bh_t - h_{p,t} \right\|^2 \right] \\
        &\leq 6 \eta^2 \LUV^2 \kappa_g E^2 + \frac{3\LUV^2}{P} \sum_{p=1}^P \E[\|\by_{t_0} - y_{p,t_0}\|^2],
    \end{align*}
    where the first inequality follows by using $\|\sum_{p=1}^P z_i\|^2 \leq P \sum_{p=1}^P  \|z_i\|^2$, the second inequality follows by the Lipschitz continuity of the gradient of $g_p$, for all $p$ by Assumptions \ref{assum:fsmooth} and \ref{assum:factorsbounded}, and the third inequality follows by Lemma \ref{lem:drift}. Combining this with the above inequality yields the first result.
    
    To see the second result, when $t \mod E = 0$,
    \begin{equation*}
        \E[\nabla g(\by_t)^T(\bh_{t+1} - \by_t)] = - \eta \E\left[\nabla g(\by_t)^T\left(\frac{1}{P} \sum_{p=1}^P G_{p,t}\right)\right].
    \end{equation*}
    Then, the proof follows by a nearly identical argument as the first case, with the replacement of $\bh_t$ by $\by_t$ and $h_{p,t}$ by $y_{p,t}$ and without applying Lemma \ref{lem:drift}.
\end{proof}

Next, we will present a general convergence result under Assumptions \ref{assum:fsmooth} and \ref{assum:factorsbounded}. In the statement of the theorem, we use the set $\mathcal{E}$ denote the set of iterations at which Algorithm \ref{alg:fedhmsimple} averages (except for the final output averaging), i.e.
\begin{equation*}
    \mathcal{E} := \{0, E, \dots, (E/T)-E\}.
\end{equation*}
We recall that by the definition of Algorithm \ref{alg:fedhmsimple}, $T \bmod E = 0$, so that $E/T$ is an integer value, so this set is well defined.

\begin{theorem} \label{thm:genericconvergence}
    Let Assumptions \ref{assum:fsmooth} and \ref{assum:factorsbounded} hold. Let $0 < \eta \leq \frac{1}{\LUV}$. Then,
    \begin{align}
        \frac{1}{T} \sum_{t=0}^{T-1} \E[\|\nabla g(\hat{h}_t)\|^2]
        &\leq \frac{2}{T \eta} \left(f(w_0)- \flow\right) + 6 \eta^2 \LUV^2 \kappa_g E^2  + \frac{\eta \LUV M_g}{P} \nonumber \\
        &\quad + \frac{2}{T \eta} \sum_{t \in \mathcal{E}} \E[g(\by_t) - g(\bh_t)] +\frac{3 \LUV^2 E}{TP} 
        \sum_{t \in \mathcal{E}} \sum_{p=1}^P \E[\|\by_{t} - y_{p,t}\|^2]. \label{eq:genericconvergence}
    \end{align}
\end{theorem}

\begin{proof}
   For any $t \in \{0,\dots, T-1\}$ such that $t \bmod E \neq 0$, by the Lipschitz continuity of the gradient of $g$,
    \begin{equation*}
        \E[g(\bh_{t+1})] \leq \E[g(\bh_t)] + \E[\nabla g(\bh_t)^T (\bh_{t+1} - \bh_t)] + \frac{\LUV}{2} \E[\|\bh_{t+1} - \bh_t\|^2].
    \end{equation*}
     Let $T_0(t)$ be the largest integer such that $T_0(t) < t$ and $T_0(t) \bmod E = 0$. Then, by Lemma \ref{lem:hbardiff}, Lemma \ref{lem:hbarinner}, the definition of $\hat{h}_t$, and $\eta \leq \frac{1}{\LUV}$, it follows that
    \begin{align}
        \E[g(\bh_{t+1})] &\leq \E[g(\bh_t)] -\frac{\eta}{2} \E[\|\nabla g(\bh_t)\|^2] + 3 \eta^3 \LUV^2 \kappa_g E^2 + \frac{\eta^2 \LUV M_g}{2 P} \nonumber \\
        &\quad- \frac{(\eta - \LUV \eta^2)}{2} \E\left[ \left\|\frac{1}{P} \sum_{p=1}^P \nabla g_p(h_{p,t})\right\|^2 \right] + \frac{3\eta \LUV^2}{2P} \sum_{p=1}^P \E[\|\by_{t_0} - y_{p,t_0}\|^2] \nonumber \\
        &\leq \E[g(\bh_t)] -\frac{\eta}{2} \E[\|\nabla g(\bh_t)\|^2] + 3 \eta^3 \LUV^2 \kappa_g E^2 + \frac{\eta^2 \LUV M_g}{2 P} \nonumber \\
        &\quad+ \frac{3\eta \LUV^2}{2P} \sum_{p=1}^P \E[\|\by_{T_0(t)} - y_{p,T_0(t)}\|^2]. \label{eq:thm1eq1}
    \end{align}
    
    On the other hand, when $t \bmod E = 0$, by the Lipschitz continuity of the gradient of $g$,
    \begin{align*}
        \E[g(\bh_{t+1})] &\leq \E[g(\by_t)] + \E[\nabla g(\by_t)^T (\bh_{t+1} - \by_t)] + \frac{\LUV}{2} \E[\|\bh_{t+1} - \by_t\|^2] \\
        &= \E[g(\bh_t)] + \E[g(\by_t) - g(\bh_t)] + \E[\nabla g(\by_t)^T (\bh_{t+1} - \by_t)] + \frac{\LUV}{2} \E[\|\bh_{t+1} - \by_t\|^2].
    \end{align*}
    Then, using Lemma \ref{lem:hbardiff}, Lemma \ref{lem:hbarinner}, the definition of $\hat{h}_t$, and $\eta \leq \frac{1}{\LUV}$,
    \begin{align}
        \E[g(\bh_{t+1})] &\leq \E[g(\bh_t)] + \E[g(\by_t) - g(\bh_t)] - \frac{\eta}{2} \E[\|\nabla g(\by_t)\|^2] + 3 \eta^3 \LUV^2 \kappa_g E^2  + \frac{\eta^2 \LUV M_g}{2 P} \nonumber \\
        &\quad- \frac{(\eta - \LUV \eta^2)}{2} \E\left[ \left\|\frac{1}{P} \sum_{p=1}^P \nabla g_p(y_{p,t})\right\|^2 \right] + \frac{\eta \LUV^2}{2P} \sum_{p=1}^P \E[\|\by_t - y_{p,t}\|^2] \nonumber \\
        &\leq \E[g(\bh_t)] + \E[g(\by_t) - g(\bh_t)] - \frac{\eta}{2} \E[\|\nabla g(\bh_t)\|^2] + 3 \eta^3 \LUV^2 \kappa_g E^2  + \frac{\eta^2 \LUV M_g}{2 P} \nonumber \\
        &\quad+ \frac{\eta \LUV^2}{2P} \sum_{p=1}^P \E[\|\by_{t} - y_{p,t}\|^2]. \label{eq:thm1eq2}
    \end{align}
    
    Now, we note that when $t \bmod E \neq 0$, it follows that $T_0(t) \in \mathcal{E}$, and for all $\hat{t} \in \{T_0(t) + 1, \dots T_0(t) + E\}$, $T_0(\hat{t}) = T_0(t)$. In addition, for any $t \in \{0, \dots, T-1\}$ such that $t \bmod E = 0$, $t \in \mathcal{E}$. Denoting the set $\{0,\dots, T-1\} \cap \{t \bmod E \neq 0\}$ by $\mathcal{E}^c$, we have
    \begin{equation*}
        \sum_{t \in \mathcal{E}^c} \sum_{p=1}^P \E[\|\by_{T_0(t)}-y_{p,T_0(t)}\|^2] + \sum_{t \in \mathcal{E}} \sum_{p=1}^P \E[\|\by_t-y_{p,t}\|^2]
        = E \sum_{t \in \mathcal{E}} \sum_{p=1}^P \E[\|\by_t-y_{p,t}\|^2].
    \end{equation*}
    
    Therefore, rearranging and summing Eq.~\eqref{eq:thm1eq1} and Eq.~\eqref{eq:thm1eq2} for all $t=0,\dots,T-1$ and using the definition of $\hat{h}_t$ in Eq.~\eqref{eq:hath},
    \begin{align*}
        \frac{\eta}{2} \sum_{t=0}^{T-1} \E[\|\nabla g(\hat{h}_t)\|^2]
        &\leq \sum_{t=0}^{T-1} \left(\E[g(\bh_t)]- \E[g(\bh_{t+1})]\right) + \sum_{t \in \mathcal{E}} \E[g(\by_t) - g(\bh_t)]\\
        &\quad  + 3 T \eta^3 \LUV^2 \kappa_g E^2  + \frac{T \eta^2 \LUV M_g}{2 P} + \frac{3\eta \LUV^2 E}{2P} \sum_{t\in \mathcal{E}} \sum_{p=1}^P \E[\|\by_{t} - y_{p,t}\|^2]
    \end{align*}
    Multiplying both sides by $\frac{2}{T\eta}$, using the facts that $g(\bh_T) = f(w(\bh_t)) \geq \flow$, and $g(\bh_0) = f(w_0)$, we have
    \begin{align*}
        \frac{1}{T} \sum_{t=0}^{T-1} \E[\|\nabla g(\hat{h}_t)\|^2]
        &\leq \frac{2}{T \eta} \left(f(w_0)- \flow\right) + 6 \eta^2 \LUV^2 \kappa_g E^2  + \frac{\eta \LUV M_g}{P}  \\
        &\quad + \frac{2}{T \eta} \sum_{t \in \mathcal{E}} \E[g(\by_t) - g(\bh_t)] +\frac{3 \LUV^2 E}{TP} 
        \sum_{t \in \mathcal{E}} \sum_{p=1}^P \E[\|\by_{t} - y_{p,t}\|^2].
    \end{align*}
\end{proof}

We provide a few remarks about Theorem \ref{thm:genericconvergence} here:
\begin{itemize}
    \item Up to constant factors, the first three terms in Eq.~\eqref{eq:genericconvergence} match those found in \citet{yu2019parallel}.
    \item The last two terms are directly due to the unique setting we consider in FedHM. The first arises due to performing the averaging step in the weight space, as opposed to the factorized space, which is where the iterates of SGD are performed. Since the iterates are averaged by first performing $W_{p,t}^{\ell} = U_{p,t}^{\ell}(V_{p,t}^{\ell})^T$ and then averaging over each $W_{p,t}^{\ell}$, the crossing terms between participants $p$ and $p'$, such as $U_{p,t}^{\ell}(V_{p',t}^{\ell})^T$, do not appear in the average. However, if one considers $\bar{W}_t^{\ell} = U_t^{\ell} (V_t^{\ell})^T$, these terms do appear, leading to the discrepancy between $g(\bar{y}_t)$ and $g(\bar{h}_t)$.
    \item The final term appears due to the heterogeneity of the devices used in the network. This term measures the average distance between a participants model after averaging versus the server's model. If all of the participants in the network used full rank models (\ie, for all $p$, $U^{\ell}_p \in \mathbb{R}^{m \times r}$, $U^{\ell}_p \in \mathbb{R}^{n \times r}$, with $r = \max\{m,n\}$), then this term would vanish.
\end{itemize}

Now, we present our main result, which follows under the addition of Assumption \ref{assum:boundedprojection}.

\begin{theorem} \label{thm:lipconvergence}
    Let Assumptions \ref{assum:fsmooth}, \ref{assum:factorsbounded}, and \ref{assum:boundedprojection} hold and let $\hat{h}_t$ be defined in Eq.~\eqref{eq:hath}. Then,
    \begin{align*}
        \frac{1}{T} \sum_{t=0}^{T-1} \E[\|\nabla g(\hat{h}_t)\|^2]
        &\leq \frac{2}{T \eta} \left(f(w_0)- \flow \right) + \frac{2L_f (L-\rho) \kappa_{\sigma_1}}{E\eta} + 6 \eta \kappa_g E\left(\eta \LUV^2 E + 2L_f\kappa_{UV}^2 \right) \nonumber  \\
        &\quad+ \frac{\eta \LUV M_g}{P} +(6 (L-\rho) \kappa_{\sigma_2}) \left(\LUV^2 + \frac{2L_f \kappa_{UV}^2}{E \eta}\right).
    \end{align*}
\end{theorem}

\begin{proof}
    To begin, by Lipschitz continuity of $f$,
    \begin{equation}
        \sum_{t \in \mathcal{E}} \E[g(\by_t) - g(\bh_t)]
        = \sum_{t \in \mathcal{E}} \E[f(w(\by_t)) - f(w(\bh_t))]
        \leq L_f \sum_{t \in \mathcal{E}} \E[\|w(\by_t) - w(\bh_t)\|^2] \label{eq:convergenceeq1}
    \end{equation}
    Then, for any $t \in \mathcal{E}$,
    \begin{align}
        \E[\|w(\by_t) - w(\bh_t)\|^2] &= \E[\|w(\by_t) - w_t + w_t - w(\bh_t)\|^2] \nonumber \\
        &\leq \E[\|w(\by_t) - w_t\|^2] + \E[\|w_t -  w(\bh_t)\|^2] \label{eq:convergenceeq2}
    \end{align}
    Next, for any $\ell \in \{\rho+1,\dots,L\}$, let $U_t^{\ell} \Sigma_t^{\ell} (V_t^{\ell})^T$ be the singular value decomposition of $w_t$, where $U_t^{\ell}$ and $V_t^{\ell}$ are orthogonal matrices and $\Sigma_t^{\ell}$ is the diagonal matrix of the singular values of $w_t^{\ell}$. Then, recalling that $\Sigma_{r(p,\ell),t}^{\ell} = \diag(\sigma_1,\dots,\sigma_{r(p,\ell)},0,\dots, 0)$, by Eq.~\eqref{eq:ypt},
    \begin{equation*}
        y_{p,t}^{\ell} = \left[ \begin{matrix} U_t^{\ell} \sqrt{\Sigma_{r(p,\ell),t}^{\ell}} \\ V_{t}^{\ell} \sqrt{\Sigma_{r(p,\ell),t}^{\ell}} \end{matrix} \right] \ \text{ and } \ \by_t^{\ell} = \left[ \begin{matrix} \frac{1}{P} \sum_{p=1}^P U_t^{\ell} \sqrt{\Sigma_{r(p,\ell),t}^{\ell}} \\ \frac{1}{P} \sum_{p=1}^P V_{t}^{\ell} \sqrt{\Sigma_{r(p,\ell),t}^{\ell}} \end{matrix}\right].
    \end{equation*}
    Denoting the vectorization of the layers $1,\dots,\rho$ by $w^{[1,\dots,\rho]}$ and recalling Eq.~\eqref{eq:splitnorm}, we have
    \begin{align}
        &\E[\|w(\by_t) - w_t\|^2] \nonumber \\
        &= \E\left[\|w(\by_t)^{[1,\dots,\rho]} - w_t^{[1,\dots,\rho]}\|^2 + \sum_{\ell=\rho+1}^{L} \|W(\by_t)^{\ell} - W_t^{\ell}\|^2_F\right] \nonumber \\
        &= \sum_{\ell=\rho+1}^{L}\E\left[ \left\|\left(\frac{1}{P} \sum_{p=1}^P U_t^{\ell} \sqrt{\Sigma_{r(p,\ell),t}^{\ell}}\right)\left(\frac{1}{P} \sum_{p=1}^P V_t^{\ell} \sqrt{\Sigma_{r(p,\ell),t}^{\ell}}\right)^T - U_t^{\ell} \Sigma_t^{\ell} (V_t^{\ell})^T\right\|^2_F\right] \nonumber \\
        &= \sum_{\ell=\rho+1}^{L}\E\left[ \left\|U_t^{\ell} \left(\left(\frac{1}{P} \sum_{p=1}^P \sqrt{\Sigma_{r(p,\ell),t}^{\ell}}\right)\left(\frac{1}{P} \sum_{p=1}^P \sqrt{\Sigma_{r(p,\ell),t}^{\ell}}\right) - \Sigma_t^{\ell}\right) (V_t^{\ell})^T\right\|^2_F\right] \nonumber \\
        &= \sum_{\ell=\rho+1}^{L} \E\left[\left\|\left(\frac{1}{P} \sum_{p=1}^P \sqrt{\Sigma_{r(p,\ell),t}^{\ell}}\right)\left(\frac{1}{P} \sum_{p=1}^P \sqrt{\Sigma_{r(p,\ell),t}^{\ell}}\right) - \Sigma_t^{\ell}\right\|^2_F\right] \nonumber \\
        &\leq \sum_{\ell=\rho+1}^{L} \kappa_{\sigma_1} = (L-\rho) \kappa_{\sigma_1}, \label{eq:convergenceeq3}
    \end{align}
    where the first equality follows by the equivalence of the Euclidean norm for vectors and Frobenius norm for matrices, the second equality follows due to  $w(\by_t)^{[1,\dots,\rho]} = w_t^{[1,\dots,\rho]}$, and the inequality follows by Assumption \ref{assum:boundedprojection}.
    
    Now, working with the second term in Eq.~\eqref{eq:convergenceeq2}, we note that $w_0 = w(\bh_0)$. Therefore, for any $t \in \mathcal{E}$ such that $t > 0$,
    \begin{align*}
        \E[\|w_t - w(\bar{h}_t)\|^2]
        &= \E\left[\|w_t^{[1,\dots,\rho]} - w(\bh_t)^{[1,\dots,\rho]}\|^2 + \sum_{\ell=\rho+1}^{L} \|W_t^{\ell} - W(\bh_t)^{\ell}\|^2_F\right] \nonumber \\
        &= \sum_{\ell=\rho+1}^{L} \E\left[\left\|\frac{1}{P} \sum_{p=1}^P U_{p,t}^{\ell} (V_{p,t}^{\ell})^T  - \left(\frac{1}{P} \sum_{i=1}^P U_{i,t}^{\ell}\right) \left( \frac{1}{P} \sum_{j=1}^P V_{j,t}^{\ell}\right)^T \right\|^2_F\right] \nonumber \\
        &= \sum_{\ell=\rho+1}^{L} \E\left[\left\|\frac{1}{P} \sum_{p=1}^P U_{p,t}^{\ell}\left( V_{p,t}^{\ell}  -\frac{1}{P} \sum_{j=1}^P V_{j,t}^{\ell}\right)^T \right\|^2_F\right] \nonumber \\
        &\leq \sum_{\ell=\rho+1}^{L} \frac{1}{P} \sum_{p=1}^P \E\left[\left\|U_{p,t}^{\ell}\left( V_{p,t}^{\ell}  -\frac{1}{P} \sum_{j=1}^P V_{j,t}^{\ell}\right)^T \right\|^2_F\right] \nonumber \\
        &\leq \kappa_{UV}^2 \sum_{\ell=\rho+1}^{L} \frac{1}{P} \sum_{p=1}^P \E\left[\left\| V_{p,t}^{\ell}  -\frac{1}{P} \sum_{j=1}^P V_{j,t}^{\ell} \right\|^2_F\right],
    \end{align*}
    where the first equality follows by the same argument as above, the second equality follows by the definitions of $w_t^{\ell}$ and $\bh_t^{\ell}$, the first inequality follows by $\|\sum_{p=1}^P Z_p\|^2_F \leq P \sum_{p=1}^P \|Z_p\|^2_F$ for any matrices $Z_p$ of the same size, and the final inequality follows by Assumption \ref{assum:boundedprojection}. Now, using the fact that $\|Z_1\|_F^2 \leq \left\|\left[\begin{matrix} Z_1\\ Z_2\end{matrix}\right]\right\|_F^2$, for any matrices $Z_1$ and $Z_2$ with the same number of columns, equation \eqref{eq:splitnorm}, and the equivalence of the Euclidean norm for vectorized matrices with the Frobenius norm, it follows that
    \begin{align}
        &\E[\|w_t - w(\bar{h}_t)\|^2] \nonumber \\
        &\leq \kappa_{UV}^2 \sum_{\ell=\rho+1}^{L} \frac{1}{P} \sum_{p=1}^P \E\left[\left\| \left[\begin{matrix} U_{p,t}^{\ell} \\ V_{p,t}^{\ell} \end{matrix} \right] - \left[\begin{matrix} \frac{1}{P} \sum_{j=1}^P U_{j,t}^{\ell} \\ \frac{1}{P} \sum_{j=1}^P V_{j,t}^{\ell} \end{matrix} \right] \right\|^2_F\right] \nonumber \\
        &\leq \kappa_{UV}^2 \frac{1}{P} \sum_{p=1}^P \E\left[\left\|h_{p,t} - \bh_t \right\|^2\right] \leq 6 \eta^2 \kappa_{UV}^2 E^2 \kappa_g + \frac{3 \kappa_{UV}^2}{P} \sum_{p=1}^P \E[\|\by_{t-E}-y_{p,t-E}\|^2], \label{eq:convergenceeq4}
    \end{align}
    where the final inequality follows by Lemma \ref{lem:drift} and by noting that $t \in \mathcal{E}$, $t > 0$ implies that $T_0(t) = t-E$. Note that it should be clear that for any $t \in \mathcal{E}$ such that $t > 0$, $t-E \in \mathcal{E}$ must hold as well. Therefore, by Theorem \ref{thm:genericconvergence}, Eq.~\eqref{eq:convergenceeq1}, Eq.~\eqref{eq:convergenceeq2}, Eq.~\eqref{eq:convergenceeq3}, Eq.~\eqref{eq:convergenceeq4}, and $|\mathcal{E}| = T/E$, we have
    \begin{align}
        \frac{1}{T} \sum_{t=0}^{T-1} \E[\|\nabla g(\hat{h}_t)\|^2]
        &\leq \frac{2}{T \eta} \left(f(w_0)- \flow\right) + 6 \eta^2 \LUV^2 \kappa_g E^2  + \frac{\eta \LUV M_g}{P} \nonumber \\
        &\quad + \frac{2L_f}{T \eta} \left(\frac{T(L-\rho) \kappa_{\sigma_1}}{E} + 6 \eta^2 \kappa_{UV}^2 T E \kappa_g + \sum_{t \in \mathcal{E}} \frac{3 \kappa_{UV}^2}{P} \sum_{p=1}^P \E[ \|\by_t-y_{p,t}\|^2]\right) \nonumber \\
        &\quad+\frac{3 \LUV^2 E}{TP} 
        \sum_{t \in \mathcal{E}} \sum_{p=1}^P \E[\|\by_{t} - y_{p,t}\|^2] \nonumber \\
        &\leq \frac{2}{T \eta} \left(f(w_0)- \flow \right) + \frac{2L_f (L-\rho) \kappa_{\sigma_1}}{E\eta} + 6 \eta \kappa_g E\left(\eta \LUV^2 E + 2L_f\kappa_{UV}^2 \right) \nonumber  \\
        &\quad+ \frac{\eta \LUV M_g}{P} +\frac{3}{TP} \left(\LUV^2 E + \frac{2L_f \kappa_{UV}^2}{\eta}\right)
        \sum_{t \in \mathcal{E}} \sum_{p=1}^P \E[\|\by_{t} - y_{p,t}\|^2]. \label{eq:convergenceeq5}
    \end{align}
    Thus, to obtain our result, we simply need to bound $\E[\|\by_t-y_{p,t}\|^2]$ for all $p$. Recalling the definitions of $\by_t$ and $y_{p,t}$, we have
    \begin{align*}
        \E[\|\by_t-y_{p,t}\|^2] &= \sum_{\ell = \rho+1}^{\ell} \E\left[\left\|\left[ \begin{matrix} \frac{1}{P} \sum_{i=1}^P U_t^{\ell} \sqrt{\Sigma_{r(i,\ell),t}^{\ell}} \\ \frac{1}{P} \sum_{i=1}^P V_t^{\ell} \sqrt{\Sigma_{r(i,\ell),t}^{\ell}} \end{matrix} \right] - \left[ \begin{matrix} U_t^{\ell} \sqrt{\Sigma_{r(p,\ell),t}^{\ell}} \\ V_t^{\ell} \sqrt{\Sigma_{r(p,\ell),t}^{\ell}} \end{matrix} \right] \right\|^2_F\right] \\
        &= \sum_{\ell = \rho+1}^{\ell} \E\left[\left\|\frac{1}{P} \sum_{i=1}^P U_t^{\ell} \sqrt{\Sigma_{r(i,\ell),t}^{\ell}}  - U_t^{\ell} \sqrt{\Sigma_{r(p,\ell),t}^{\ell}} \right\|^2_F\right] \\
        &\quad+ \E\left[\left\|\frac{1}{P} \sum_{i=1}^P V_t^{\ell} \sqrt{\Sigma_{r(i,\ell),t}^{\ell}}  - V_t^{\ell} \sqrt{\Sigma_{r(p,\ell),t}^{\ell}} \right\|^2_F\right] \\
        &= \sum_{\ell = \rho+1}^{\ell} \E\left[\left\|U_t^{\ell} \left(\frac{1}{P} \sum_{i=1}^P \sqrt{\Sigma_{r(i,\ell),t}^{\ell}}  - \sqrt{\Sigma_{r(p,\ell),t}^{\ell}}\right) \right\|^2_F\right] \\
        &\quad+ \E\left[\left\|V_t^{\ell} \left(\frac{1}{P}\sum_{i=1}^P \sqrt{\Sigma_{r(i,\ell),t}^{\ell}}  -  \sqrt{\Sigma_{r(p,\ell),t}^{\ell}}\right) \right\|^2_F\right] \\
        &= 2\sum_{\ell = \rho+1}^{\ell} \E\left[\left\|\frac{1}{P} \sum_{i=1}^P \sqrt{\Sigma_{r(i,\ell),t}^{\ell}}  - \sqrt{\Sigma_{r(p,\ell),t}^{\ell}}\right\|^2_F\right] \\
        &\leq 2 (L-\rho) \kappa_{\sigma_2},
    \end{align*}
    where the last equality follows by the orthogonality of $U_t^{\ell}$ and $V_t^{\ell}$ and the inequality follows by Assumption \ref{assum:boundedprojection}. Combining this inequality with Eq.~\eqref{eq:convergenceeq5} and recalling that $|\mathcal{E}| = T/E$ yields the result.
\end{proof}

We remark here that to obtain the order notation result in Theorem \ref{thm:convergencemain}, we let $\kappa_{\sigma} = \max\{\kappa_{\sigma_1},\kappa_{\sigma_2}\}$ and note that due to the definitions of $M_g$ and $\kappa_g$, $M_g = \mathcal{O}(\kappa_g)$. Therefore, it follows directly from Theorem \ref{thm:lipconvergence} that
\begin{equation*}
    \frac{1}{T} \sum_{i=0}^{T-1} \E[\|\nabla g(\hat{h}_i)\|^2] \leq \OC\left(\frac{f(w_0) - \flow}{T  \eta} + \frac{\kappa_g \eta }{P} + \kappa_g \max\{E \eta, E^2 \eta^2\} + \kappa_{\sigma}\left(1 + \frac{1}{E \eta}\right) \right),
\end{equation*}
which matches the result of Theorem \ref{thm:convergencemain} after the redefinition of $T$ in Algorithm \ref{alg:fedhmsimple}.

\section{Detailed Experimental Setup}
\label{appendx:c}
In this section, we elaborate the details of experiments, including datasets and hyper-parameters. Our implementation is based on PyTorch and we use the MPI as the communication backend. All the experiments are conducted on a 256GB server with 4$\times$Tesla V100 GPUs in parallel.

\subsection{Datasets}
\noindent \textbf{CIFAR-10} CIFAR-10~\cite{krizhevsky2009learning} is the popular classification benchmark dataset, which 
consists of 60,000 images with the resolution of $32\times32$, covering 10 classes, with 6,000 images per class.

\noindent \textbf{CIFAR-100} 
Similar to CIFAR-10, CIFAR-100~\cite{krizhevsky2009learning} is a image dataset of 100 classes, with 600 images per class. 

\noindent \textbf{Tiny-ImageNet} Tiny-ImageNet contains 100,000 images of 200 classes (500 for each class) downsized to $64\times64$ colored images. Each class has 500 training images, 50 validation images and 50 test images.

\subsection{Detailed Implementation in experiments}
\noindent\textbf{Computation Complexity Setting}. For \system, we choose the $\gamma$ from $\{0.5, 0.25, 0.125\}$ on ResNet-18 and ResNet-34; set the $\rho=2$ in ResNet-18 and $\rho=15$ in ResNet-34. Table~\ref{tbl:modified_resnet18} and Table~\ref{tbl:modified_resnet34} shows the detailed architecture. For HeteroFL, we set the shrinkage ratio of HeteroFL in
$\{0.64, 0.50, 0.40,0.35\}$ on ResNet-18 and ResNet-34, so as to reach a larger model size.  For Split-Mix, we control the smallest model the same size as HeteroFL, and tune the number of base models in $\{1,2,3,7,9\}$.

\noindent\textbf{Detailed Implementation of \system}. The softmax temperature $\tau$ is used in dynamic heterogeneous setting, since all the models are trained on every clients, the larger model should be given a higher weight. Following~\cite{diao2020heterofl}, we incorporate the static batch normalization(sBN) and masking cross-entropy methods into all the methods in the experiments. Note that to accelerate the training process, we use sBN when the communication rounds are greater than the first milestone round.

\noindent\textbf{Federated Learning Settings}.
Table~\ref{tbl:experiment_setting} presents the detailed experimental settings in our experiments. 
Note that we use MultiStepLR as scheduler, and the decay rate is set to $0.1$.

\begin{table}[!htb]
\centering
\caption{Detailed experimental settings in our experiments.}
\label{tbl:experiment_setting}
\resizebox{.75\linewidth}{!}{
\begin{tabular}{lccc}
\hline
Dataset             & CIFAR-10      & CIFAR-100   & Tiny-ImageNet \\
Clients       & \multicolumn{3}{c}{20}                      \\
\#Samples/client     & 2,500          & 2,500        & 5,000          \\
Client Sampling Rate    & \multicolumn{3}{c}{0.5}                     \\
Local Epoch         & \multicolumn{3}{c}{10}                      \\
Batch Size          & \multicolumn{3}{c}{64}                      \\
Optimizer           & \multicolumn{3}{c}{SGD}                     \\
Momentum            & \multicolumn{3}{c}{0.9}                     \\
Weight decay        & \multicolumn{3}{c}{$1e\text{-}4$}                    \\
Learning Rate       & \multicolumn{3}{c}{0.1}                     \\
Model               & ResNet-18  & ResNet-34    & ResNet-34 \\
Communication Round & 160           & 100         & 60            \\
Milestone Round& {[}100, 150{]} & {[}70, 90{]} & {[}40, 55{]}   \\ \hline
\end{tabular}
}

\end{table}

\subsection{Hybrid Network Structure}
\begin{figure}[tb]
\centering
\begin{subfigure}{0.485\columnwidth}
\centering
\includegraphics[width=\linewidth]{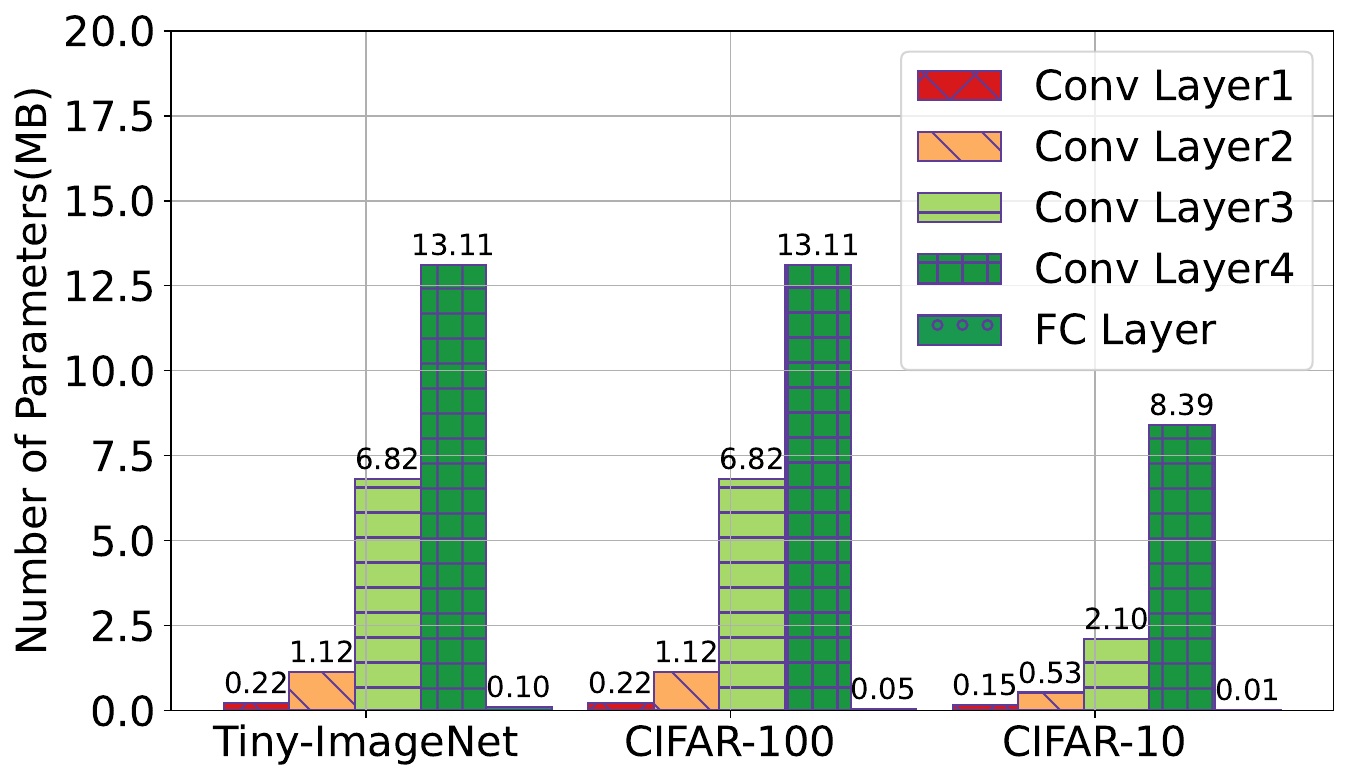}
\caption{Parameters of different layers on different datasets.}
\label{fig:params_analysis}
\end{subfigure}
\begin{subfigure}{0.485\columnwidth}
\centering
\includegraphics[width=\linewidth]{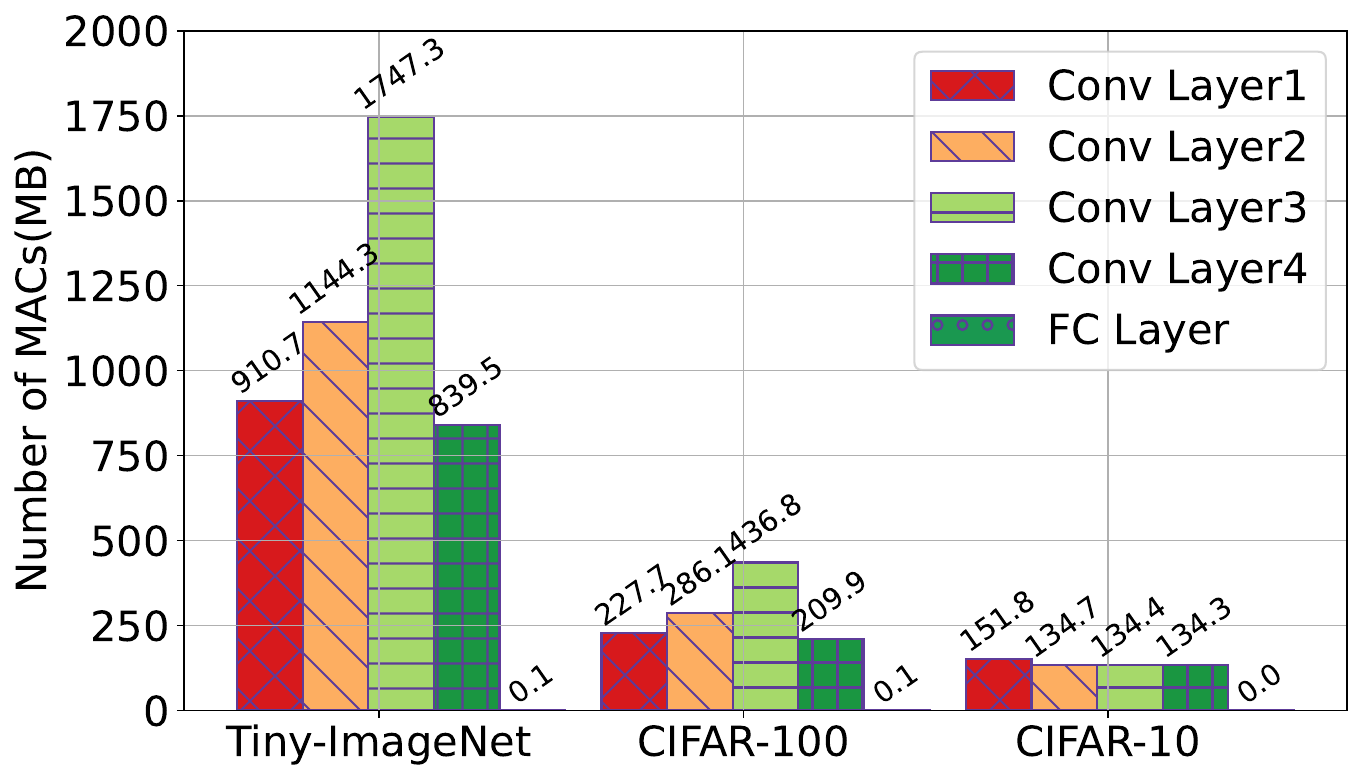}
\caption{MACs of different layers on different datasets.}
\label{fig:macs_analysis}
\end{subfigure}
\caption{Parameters and MACs of different layers of ResNet on various datasets. FC Layer denotes Fully-Connected Layer.}
\label{fig:network_analysis}
\end{figure}

For CIFAR-10, CIFAR-100, and Tiny-ImageNet datasets, we modify the ResNet architecture used on ImageNet,
and the modified network structures are shown in Tables~\ref{tbl:modified_resnet18} and \ref{tbl:modified_resnet34}. To measure the model size and computation complexity, we also analyze the number of  parameters and the number of multiply–accumulate operations (MACs) of both ResNet-18 and ResNet-34 on different datasets, as shown in Figure~\ref{fig:network_analysis}. 
We find that the later layers (Layer3 and Layer4) contribute the most parameters, \ie, over 93\% for ResNet-34 and ResNet-18, the two later layers also contribute 55.6\% computation overhead on ResNet-34. From Figure~\ref{fig:network_analysis}, we can conclude
that compressing the later layers can significantly reduce the communication cost and computation overhead.

\begin{table}[tb]
\centering
\caption{The hybrid ResNet-18 architecture for CIFAR-10 dataset in the experiments.}
\label{tbl:modified_resnet18}
\renewcommand\arraystretch{1.0}
\resizebox{.95\linewidth}{!}{
\begin{tabular}{lcc}
\hline
Layer Name & ResNet-18 & Low Rank ResNet-18 \\
conv1      & 3$\times$3, 64, stride 1, padding 1&  3$\times$3, 64, stride 1, padding 1 \\
layer1     & $\left[\begin{array}{cc}
     3\times3, 64\\
     3\times3, 64\end{array}  \right] \times2$&  $\left[\begin{array}{cc}
     3\times3, 64\\
     3\times3, 64
\end{array}  \right], 
\left[\begin{array}{cc}
     conv_u(64, 64*r, 3,1)\\
     conv_v(64*r, 64, 1,3) 
\end{array}  \right]\times2$                  \\
layer2 &  $\left[\begin{array}{cc}
     3\times3, 128\\
     3\times3, 128\end{array}  \right] \times2$ &  
     $\left[\begin{array}{cc}
     conv_u(128, 128*r, 3,1)\\
     conv_v(128*r, 128, 1,3) 
\end{array}  \right] \times4$ \\ 
layer3 &  $\left[\begin{array}{cc}
     3\times3, 256\\
     3\times3, 256\end{array}  \right] \times2$ &  
     $\left[\begin{array}{cc}
     conv_u(256, 256*r, 3,1)\\
     conv_v(256*r, 256, 1,3) 
\end{array}  \right] \times4$ \\ 
layer4 & $\left[\begin{array}{cc}
     3\times3, 512\\
     3\times3, 512\end{array}  \right] \times2$ &  
     $\left[\begin{array}{cc}
     conv_u(512, 512*r, 3,1)\\
     conv_v(512*r, 512, 1,3) 
\end{array}  \right] \times4$ \\ 
\hline
\end{tabular}
}
\end{table}

\begin{table}[tb]
\centering
\caption{The hybrid ResNet-34 architecture for CIFAR-100 and Tiny-ImageNet datasets in the experiments.}
\label{tbl:modified_resnet34}
\renewcommand\arraystretch{1.0}
\resizebox{.85\linewidth}{!}{
\begin{tabular}{lcc}
\hline
Layer Name & ResNet-34 & Low Rank ResNet-34 \\
conv1      & 3$\times$3, 64, stride 1, padding 1&  3$\times$3, 64, stride 1, padding 1 \\
layer1     & $\left[\begin{array}{cc}
     3\times3, 64\\
     3\times3, 64\end{array}  \right] \times3$&  $\left[\begin{array}{cc}
     3\times3, 64\\
     3\times3, 64
\end{array}  \right]\times3$      \\ 

layer2 &  $\left[\begin{array}{cc}
     3\times3, 128\\
     3\times3, 128\end{array}  \right] \times4$ &  
    $\left[\begin{array}{cc}
     3\times3, 128\\
     3\times3, 128\end{array}  \right] \times4$ \\ 

layer3 &  $\left[\begin{array}{cc}
     3\times3, 256\\
     3\times3, 256\end{array}  \right] \times6$ &  
     $\left[\begin{array}{cc}
     conv_u(256, 256*r, 3,1)\\
     conv_v(256*r, 256, 1,3) 
\end{array}  \right] \times12$ \\ 
layer4 & $\left[\begin{array}{cc}
     3\times3, 512\\
     3\times3, 512\end{array}  \right] \times$3 &  
     $\left[\begin{array}{cc}
     conv_u(512, 512*r, 3,1)\\
     conv_v(512*r, 512, 1,3) 
\end{array}  \right] \times6$ \\ 
\hline
\end{tabular}
}

\end{table}

\section{Additional Experiments}

\subsection{Ablation Study}

\begin{figure}[htb]
\centering
\begin{subfigure}{0.48\columnwidth}
\includegraphics[width=0.9\linewidth]{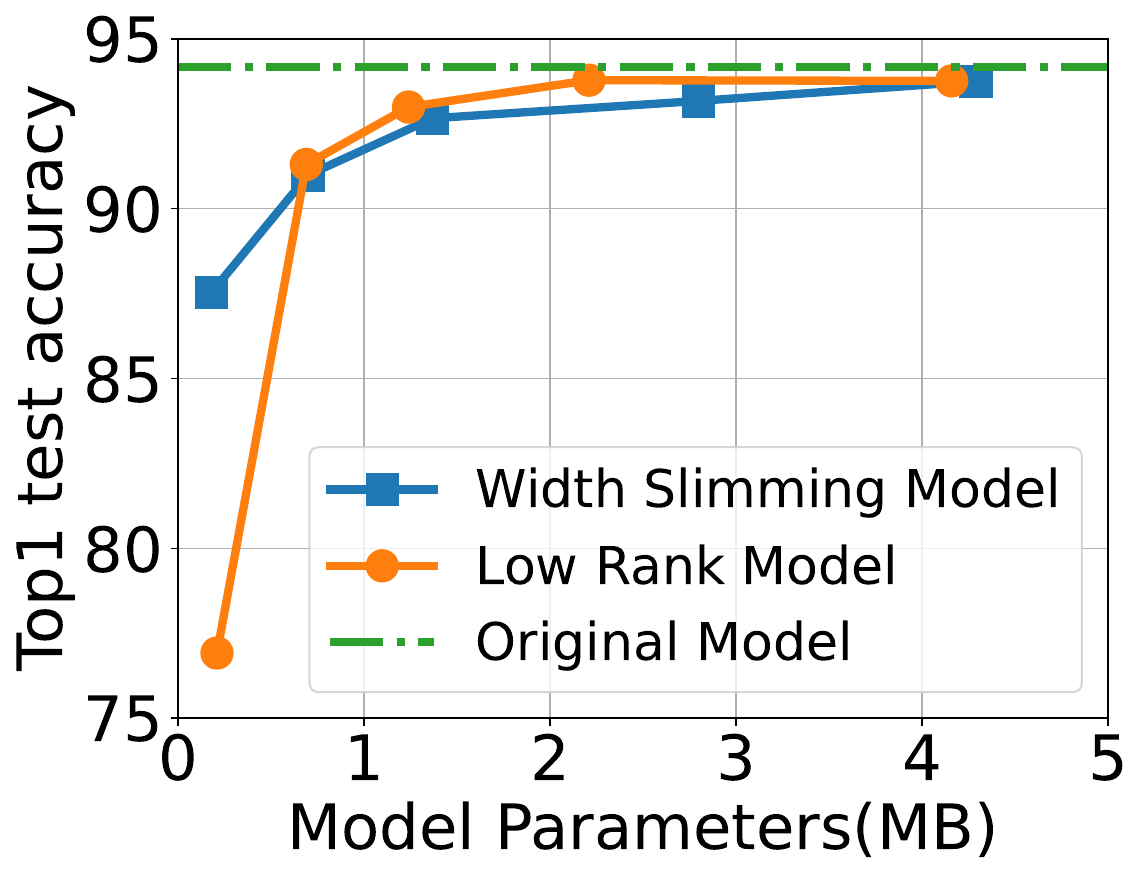}
\caption{CIFAR-10.}
\end{subfigure}
\begin{subfigure}{0.48\columnwidth}
\includegraphics[width=0.9\linewidth]{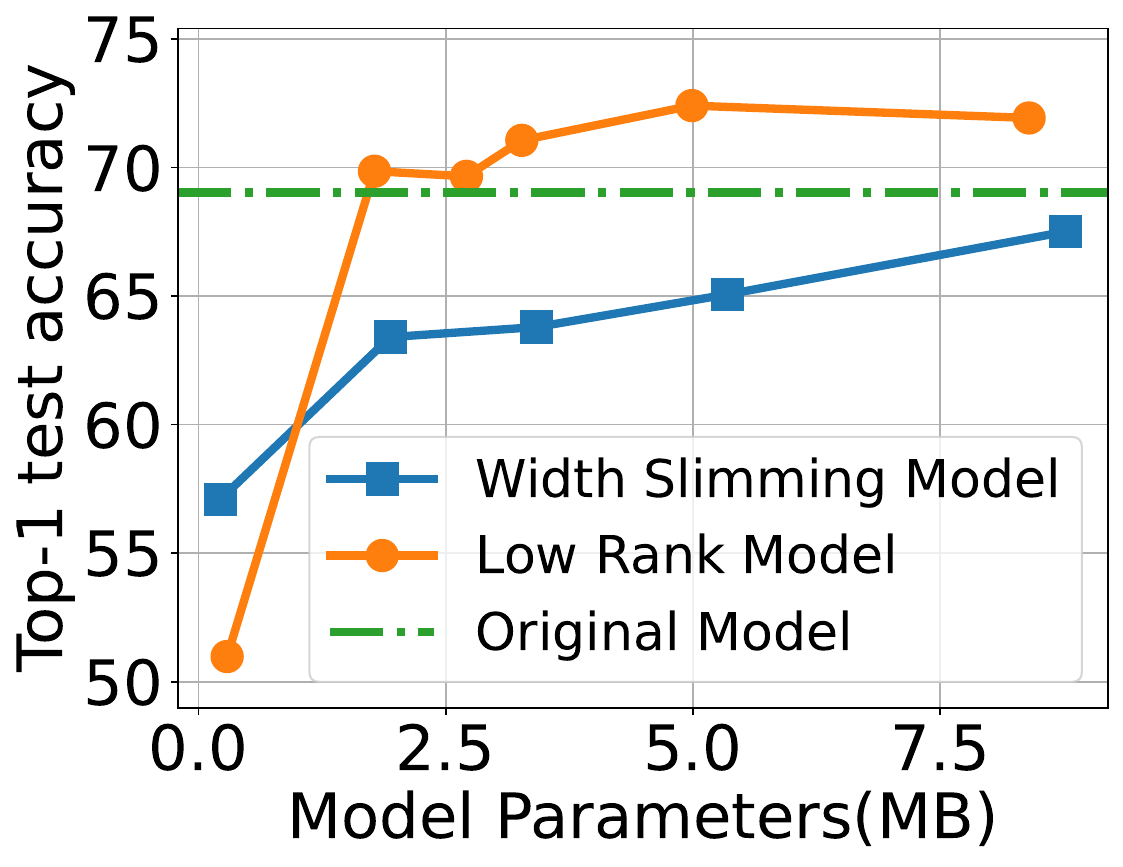}
\caption{CIFAR-100.}
\label{fig:cifar100_param}
\end{subfigure}
\caption{Comparison of different compression methods in terms of Top-1 test accuracy.}
\label{fig:lowrank_param}

\end{figure}

\noindent\textbf{Effect of Compression.}
To study the effect of different compression methods, we conduct experiments on training low-rank models and width slimming models in homogeneous setting respectively. 
For the low-rank model with the smallest size, we remove the hybrid network structure and directly decompose all the convolution layers. 
As shown in Figure~\ref{fig:lowrank_param}, if not compressed severely(parameters ratio $\geq 5\%$), 
low-rank models consistently outperform width slimming model with fewer parameters. However, when removing the hybrid network structure and setting the model size as 2\% (the rank of each convolution layer is $k$), the performance of vanilla low-rank model will drop significantly,
\ie, from over 90\% to below 80\% on CIFAR-10, and from 70\% to below 55\% on CIFAR-100, indicating
the effectiveness of hybrid network structure.
From Figure~\ref{fig:cifar100_param}, we can also find that low-rank models achieve comparable performance on CIFAR-10, and better performance on CIFAR-100, when compared with the original model.
This is because that SI and FD can improve the performance of low-rank models, and the model compression can lead to regularization effects.  

\noindent\textbf{Aggregation of Different Models.}
We conduct experiments on different low-rank models combination and extend two variants based on \system (named \system-V2 and \system-V3), under fixed heterogeneous setting, to study the effect of aggregation. 
As shown in Table~\ref{tbl:eval_effect_fix}, \system is robust to different models aggregation, and the performance of small model increases as aggregating with larger models. 
The experiment results show that the aggregation method in \system will not hurt the performance of small model under fixed heterogeneous setting.
\begin{table}[!hb]
\caption{Evaluating the effectiveness with different factorized model combination on CIFAR-10, under fixed heterogeneous setting.}
\label{tbl:eval_effect_fix}
\centering
\resizebox{0.98\linewidth}{!}{%
\begin{tabular}{l|cccccccc}  %
\hline
\multirow{2}{*}{Setting} & \multicolumn{2}{c}{\system} & \multicolumn{2}{c}{\system-V2} & \multicolumn{2}{c}{\system-V3} & \multicolumn{2}{l}{Low Rank Model}  \\ 
\cline{2-9} 
    & Parameters   & Accuracy   & Parameters     & Accuracy    & Parameters     & Accuracy    & \multicolumn{1}{c}{Parameters} & \multicolumn{1}{c}{Accuracy} \\ 
 \hline
  \multirow{4}{*}{IID}     & 11.17M       & \bf{93.50}      & 4.16M($\times$2)          & 93.33       & 2.21M($\times$3)          & 93.27       & 1.24M($\times$4)                          & 92.99                        \\
                          & 4.16M        & \bf{93.52}      & 2.21M          & 93.36       & 1.24M          & 93.19       &                                &                              \\
                          & 2.21M        & \bf{93.49}      & 1.24M          & 93.35       &                &             &                                &                              \\
                          & 1.24M        & \bf{93.43}      &                &             &                &             &                                &                              \\
\hline
\multirow{4}{*}{Non-IID} & 11.17M       & \bf{91.59}      & 4.16M($\times$2)          & 91.18       & 2.21M($\times$3)          & 90.93       & 1.24M($\times$4) & 90.95               \\
                      & 4.16M        & \bf{91.51}      & 2.21M          & 91.17       & 1.24M          & 90.95       &                                &                             \\
                      & 2.21M        & \bf{91.55}      & 1.24M          & 91.25       &                &             &                                &                              \\
                      & 1.24M        & \bf{91.47}      &                &             &                &             &                                &                              \\ 
\hline
\end{tabular}
}
\end{table}
\subsection{Performance under Homogeneous Setting}

\begin{table*}[t] 
\centering
\caption{Evaluating different methods in terms of Top-1 test accuracy under homogeneous settings.}
\label{tbl:homo}
\renewcommand\arraystretch{1.0}
\resizebox{0.9\linewidth}{!}{
\begin{tabular}{l|l|cccc}
\hline
\multirow{2}{*}{DataSet}& \multirow{2}{*}{Setting} & \multicolumn{2}{c}{Low Rank Model} & \multicolumn{2}{c}{Width Slimming Model} \\
        \cline{3-6}
      & & Parameters  & Accuracy          & Parameters  & Accuracy         \\
\hline

\multirow{8}{*}{CIFAR-10}& \multirow{4}{*}{IID}     & 11.17M  & 94.16 & 11.17M &  94.16  \\
                 &  & \bf{4.16M }      & \bf{93.76}        & 4.29M    & 93.75     \\
                 & & \bf{2.21M}       & \bf{93.78}               & 2.80M         & 93.17    \\
                 &  & \bf{1.24M}      & \bf{92.99}           & 1.37M          & 92.66       \\
\cline{2-6}
& \multirow{4}{*}{Non-IID} & 11.17M     & 92.50        & 11.17M         &  92.50   \\
             &  & \bf{4.16M}                 & \bf{92.25}            & 4.29M          &  92.08   \\
             & & \bf{2.21M}                 & \bf{92.05}                 & 2.80M     & 91.25   \\
             & & \bf{1.24M}                 & \bf{90.95}                   & 1.37M   & 90.21 \\  
\hline
\multirow{8}{*}{CIFAR-100} & \multirow{4}{*}{IID}  
        &    21.33M     &      69.02   & 21.33M         & 69.02    \\
         & & \bf{8.40M}   & \bf{71.92} & 8.77M       & 67.50      \\
         & & \bf{4.99M}    & \bf{72.40}  & 5.35M       & 65.06       \\
         & & \bf{3.27M}  & \bf{71.05}  & 3.42M          & 63.07    \\
\cline{2-6}
& \multirow{4}{*}{Non-IID} & 21.33M     &  68.13  & 21.33M  &  68.13      \\
      &  & \bf{8.40M}    & \bf{70.69}     & 8.77M          & 68.32     \\
      & & \bf{4.99M}  &  \bf{69.70}   & 5.35M          &  66.73     \\
      & & \bf{3.27M} & \bf{67.96}   & 3.42M          & 66.69   \\ 
\hline
\multirow{8}{*}{Tiny-ImageNet} & \multirow{4}{*}{IID} 
         & 21.38M  &56.20 & 21.38M  &56.20   \\
         & &\bf{8.45M} & \bf{58.68}  & 8.80M  & 53.19      \\
         & &\bf{5.04M}  & \bf{57.27} & 5.38M  & 52.26      \\
         & &\bf{3.33M} & \bf{55.92}  & 3.44M  & 50.89       \\
\cline{2-6}
& \multirow{4}{*}{Non-IID} &  21.38M & 56.00 & 21.38M & 56.00 \\
        & &\bf{8.45M} & \bf{55.29}  & 8.80M  & 53.61    \\
       & & \bf{5.04M}  & \bf{53.07}  & 5.38M  & 52.80     \\
         & &\bf{3.33M} & \bf{50.37} & 3.44M  & 50.19      \\
\hline
\end{tabular} 
}

\end{table*}

In Table~\ref{tbl:homo}, we evaluate different compression methods under homogeneous FL setting in terms of Top-1 test accuracy. It clearly shows that our proposed compression method outperforms the uniform pruning method in terms of accuracy, with a smaller model size. 
Comparing with the original model, factorization compression can significantly reduce the communication cost by 7.8$\times$ to 9.0$\times$ and decrease the computation overhead by 2.0$\times$. 
Note that on CIFAR-100, we directly compare the low-rank model (8.40M) with the original model (21.33M) and find that the low-rank model outperforms the original model. As shown in Table~\ref{tbl:homo}, under IID and Non-IID settings, training the smallest low-rank model (3.33M) homogeneously achieves 55.92\% and 50.37\% test accuracy on Tiny-ImageNet.

\end{document}